\documentclass[journal]{IEEEtran}
\usepackage{amsmath,graphicx}
\usepackage[T1]{fontenc}
\usepackage{hyperref, times, cite}
\usepackage{url, datetime}
\usepackage{mathtools}
\usepackage{bm, bbm}
\usepackage{cool}
\usepackage{amssymb,amsfonts,amsmath,amsthm,amscd,dsfont,mathrsfs}
\usepackage{graphicx,float,psfrag,epsfig,amssymb}
\usepackage{wrapfig}
\usepackage{relsize}
\usepackage{color}
\usepackage{pict2e}
\usepackage{algorithm}
\usepackage{algcompatible}
\usepackage{algorithmicx}
\usepackage{caption}
\usepackage{nameref}
\usepackage{enumerate}
\usepackage{stackrel}

\usepackage{caption}
\usepackage{subcaption}

\usepackage{amsthm}

\begin{document}
\renewcommand{\algorithmicrequire}{\textbf{Input:}}
\renewcommand{\algorithmicensure}{\textbf{Output:}}
\thispagestyle{empty}

\title{Learning High-Dimensional Differential Graphs From Multi-Attribute Data}
\author{ Jitendra K.\ Tugnait 
\thanks{J.K.\ Tugnait is with the Department of 
Electrical \& Computer Engineering,
200 Broun Hall, Auburn University, Auburn, AL 36849, USA. 
Email: tugnajk@auburn.edu . }

\thanks{This work was supported by the National Science Foundation Grants ECCS-2040536 and CCF-2308473.}}

\maketitle

\renewcommand{\algorithmicrequire}{\textbf{Input:}}
\renewcommand{\algorithmicensure}{\textbf{Output:}}

\begin{abstract}
We consider the problem of estimating differences in two Gaussian graphical models (GGMs) which are known to have similar structure. The GGM structure is encoded in its precision (inverse covariance) matrix. In many applications one is interested in estimating the difference in two precision matrices to characterize underlying changes in conditional dependencies of two sets of data. Existing methods for differential graph estimation are based on single-attribute (SA) models where one associates a scalar random variable with each node. In multi-attribute (MA) graphical models, each node represents a random vector. In this paper, we analyze a group lasso penalized D-trace loss function approach for differential graph learning from multi-attribute data. An alternating direction method of multipliers (ADMM) algorithm is presented to optimize the objective function. Theoretical analysis establishing consistency in support recovery and estimation in high-dimensional settings is provided. Numerical results based on synthetic as well as real data are presented.
\end{abstract}

\vspace*{-0.1in}

\section{Introduction} \label{intro}

\IEEEPARstart{G}{raphical} models provide a powerful tool for analyzing multivariate data \cite{Lauritzen1996, Whittaker1990}. In a statistical graphical model, the conditional statistical dependency structure among $p$ random variables $x_1, x_1, \cdots , x_p$, is represented using an undirected graph ${\cal G} = \left( V, {\cal E} \right)$, where $V = \{1,2, \cdots , p\} =[p]$ is the set of $p$ nodes corresponding to the $p$ random variables $x_i$s, and ${\cal E} \subseteq V \times V$ is the set of undirected edges describing conditional dependencies among the components of ${\bm x}$. The graph ${\cal G}$ then is a conditional independence graph (CIG) where there is no edge between nodes $i$ and $j$ (i.e., $\{i,j\} \not\in {\cal E}$) iff $x_i$ and $x_j$ are conditionally independent given the remaining $p$-$2$ variables $x_\ell$, $\ell \in [p]$, $\ell \neq i$, $\ell \neq j$. In particular, Gaussian graphical models (GGMs) are CIGs where ${\bm x}$ is multivariate Gaussian. Suppose ${\bm x}$ has positive-definite covariance matrix $\bm{\bm \Sigma}$ with inverse covariance matrix $\bm{\Omega} = \bm{\bm \Sigma}^{-1}$. Then ${\Omega}_{ij}$, the $(i,j)$-th element of  $\bm{\Omega}$, is zero iff $x_i$ and $x_j$ are conditionally independent. Such models for ${\bm x}$ have been extensively studied. Given $n$ samples of ${\bm x}$, in {\em high-dimensional settings} where $p \gg 1$ and/or $n$ is of the order of $p$, one estimates $\bm{\Omega}$ under some sparsity constraints; see \cite{Danaher2014, Meinshausen2006, Mohan2014, Friedman2008}. 

More recently there has been increasing interest in differential network analysis where one is interested in estimating the difference in two inverse covariance matrices \cite{Yuan2017, Tang2020, Wu2020}. Given observations ${\bm x}$ and ${\bm y}$ from two groups of subjects, one is interested in the difference ${\bm \Delta} = {\bm \Omega}_y - {\bm \Omega}_x$, where ${\bm \Omega}_x = (E \{ {\bm x} {\bm x}^\top \} )^{-1}$ and ${\bm \Omega}_y = (E \{ {\bm y} {\bm y}^\top \} )^{-1}$. The associated differential graph is ${\cal G}_\Delta = \left( V, {\cal E}_\Delta \right)$ where $\{i,j\} \in {\cal E}_\Delta$ iff ${\bm \Delta}_{ij} \ne 0$. It characterizes differences between the GGMs of the two sets of data. We use the term differential graph as in \cite{Zhao2019, Zhao2022} (\cite{Yuan2017, Tang2020, Wu2020} use the term differential network). As noted in \cite{Tang2020}, in biostatistics, the differential network/graph describes the changes in conditional dependencies between components under different environmental or genetic conditions. For instance, one may be interested in the differences in the graphical models of healthy and impaired subjects, or models under different disease states, given gene expression data or functional MRI signals \cite{Danaher2014, Zhao2014, Belilovsky2016}. 

In the preceding graphs, each node represents a scalar random variable.  In many applications, there may be more than one random variable associated with a node. This class of graphical models has been called multi-attribute (MA) graphical models in \cite{Kolar2013, Kolar2014, Chiquet2018, Tugnait21a} and vector graphs or networks in \cite{Marjanovic18, Yue20, Sundaram20, Yue21}. In a gene regulatory network, one may have different molecular profiles available for a single gene, such as protein, DNA and RNA. Since these molecular profiles are on the same set of biological samples, they constitute multi-attribute data for gene regulatory graphical models in \cite{Kolar2014, Chiquet2018}. Consider $p$ jointly Gaussian vectors ${\bm z}_i \in \mathbb{R}^m$, $i \in [p]$. We associate ${\bm z}_i$ with the $i$th node of graph ${\cal G} = \left( V, {\cal E} \right)$,  $V = [p]$, ${\cal E} \subseteq V \times V$. We now have $m$ attributes per node. Now $\{ i,j \} \in {\cal E}$ iff vectors ${\bm z}_i$ and ${\bm z}_j$ are conditionally independent given the remaining $p$-$2$ vectors $\{ {\bm z}_\ell \, , \ell \in V \textbackslash \{i_, j \} \}$. Let ${\bm x} = [ {\bm z}_1^\top \; {\bm z}_2^\top \; \cdots \; {\bm z}_p^\top ]^\top \in \mathbb{R}^{mp}$. 
Let ${\bm \Omega} = (E \{ {\bm x} {\bm x}^\top \} )^{-1}$ assuming $E \{ {\bm x} {\bm x}^\top \} \succ {\bm 0}$. Define the $m \times m$ subblock ${\bm \Omega}^{(ij)}$ of ${\bm \Omega}$ as $[{\bm \Omega}^{(ij)}]_{rs} = [{\bm \Omega}]_{(i-1)m+r, (j-1)m+s} \, , \; r,s=1,2, \cdots , m$.  Then we have the following equivalence \cite[Sec.\ 2.1]{Kolar2014}
\begin{equation}  \label{neweq24}
  \{ i,j \} \not\in {\cal E} \; \Leftrightarrow \; {\bm \Omega}^{(ij)} = {\bm 0} \, .
\end{equation}

This paper is concerned with estimation of differential graphs from multi-attribute data. Given independent and identically distributed (i.i.d.) samples ${\bm x}(t)$, $t=1,2, \cdots , n_x$, of ${\bm x} = [ {\bm z}_1^\top \; {\bm z}_2^\top \; \cdots \; {\bm z}_p^\top ]^\top \in \mathbb{R}^{mp}$ where ${\bm z}_i \in \mathbb{R}^m$, $i \in [p]$, are jointly Gaussian, and similarly given samples ${\bm y}(t)$, $t=1,2, \cdots , n_y$, of ${\bm y} \in \mathbb{R}^{mp}$, our objective is to estimate the difference ${\bm \Delta} = {\bm \Omega}_y - {\bm \Omega}_x$, and determine the differential graph ${\cal G}_\Delta = \left( V, {\cal E}_\Delta \right)$ with edgeset ${\cal E}_\Delta = \{ \{ k, \ell \} \,:\, \| {\bm \Delta}^{(k \ell)} \|_F \ne 0 \}$.
\vspace*{-0.1in}

\subsection{Related Work} 
All prior work on high-dimensional differential graph estimation from i.i.d.\ samples addresses single-attribute (SA) models where each node represents a scalar random variable. One naive approach would be to estimate the two precision matrices separately by any existing estimator (see \cite{Meinshausen2006, Friedman2008} and references therein) and then calculate their difference to estimate the differential graph. (This approach is also applicable to MA graphs.) This approach estimates twice the number of parameters, hence needs larger sample sizes for same accuracy, and also imposes sparsity constraints on each precision matrix for the methods to work. The same comment applies to methods such as \cite{Danaher2014, Mohan2014}, where the two precision matrices and their differences are jointly estimated. A recent survey is in \cite{Tsai2022}. In these approaches, given $K \ge 2$ related groups of data, each $p$-variate and sharing the same set of nodes $V$, but possibly differing in connected edgesets, the objective is to jointly estimate the $K$ precision matrices and their pairwise differences, with sparsity constraints on each of the $K$ precision matrices and their pairwise differences.  These approaches require each of the $K$ precision matrices to be sparse. If only the differences in the precision matrices is of interest, alternative approaches exist where no sparsity constraints are imposed on individual precision matrices. For instance, direct estimation of the difference in the two precision matrices has been considered for SA graphs in \cite{Zhang2009, Zhao2014, Liu2014, Xu16, Yuan2017, Liu2017a, Liu2017, Jiang2018, Tang2020, Wu2020}, where only the difference is required to be sparse, not the two individual precision matrices. In \cite{Zhang2009, Xu16, Yuan2017, Jiang2018, Tang2020, Wu2020} precision difference matrix estimators are based on  a D-trace loss \cite{Zhang2014}, while \cite{Zhao2014} discusses a Dantzig selector type estimator. In \cite{Liu2014, Liu2017a, Liu2017} differential graph is estimated by directly modeling the ratio of the probability densities of the random vectors under the two graphs. 

Estimation of MA differential graphs has not been investigated before.  The work of \cite{Zhao2019, Zhao2022} is similar to an MA formulation except that in \cite{Zhao2019, Zhao2022}, ${\bm x}(t)$ and ${\bm y}(t)$ are non-stationary (``functional'' modeling), and instead of a single record (sample) of ${\bm x}(t)$, $t=1,2, \cdots , n_x$ and ${\bm y}(t)$, $t=1,2, \cdots , n_y$, as in this paper, they assume multiple independent observations of ${\bm x}(t)$, $t \in {\cal T}$ (a closed subset of real line), and ${\bm y}(t)$, $t \in {\cal T}$. The objective function in \cite[Eqns.\ (10)-(11)]{Zhao2022} is the same as our objective function (\ref{eqn15})-(\ref{eqn20}), but consequent estimation of edges and theoretical analysis are vastly different. We estimate edges as in (\ref{eqn26}), i.e., our threshold is set at zero and this is the method analyzed in our Theorem 1(iv) for graph recovery with high probability. In \cite{Zhao2019, Zhao2022}, this threshold is set at a parameter $\epsilon_n > 0$ (see \cite[Eqn.\ (13)]{Zhao2022}) which is a function of sample size $n$, number of nonzero entries in true ${\bm \Delta}$, smallest eigenvalues of true covariances ${\bm \Omega}_y^{-1}$ and ${\bm \Omega}_x^{-1}$ (in our notation), and several other factors. That is, $\epsilon_n$ is unknowable for practical implementation and it is used as a theoretical construct to establish graph support recovery in \cite[Theorem 10]{Zhao2022}. In simulations, \cite{Zhao2019, Zhao2022}  set $\epsilon_n =0$. That is, \cite{Zhao2019, Zhao2022} do not analyze what they implement (the proof does not hold for $\epsilon_n =0$), and they do not implement what they analyze ($\epsilon_n$ is unknowable). There is no counterpart to our Theorem 1 in \cite{Zhao2019, Zhao2022}, and the methodology of our Theorem 1 allows us to set the edge detection threshold to zero. Our Theorem 2 follows the general framework of \cite{Negahban2012} to bound the Frobenius norm of the error in estimating ${\bm \Delta}$, and \cite{Zhao2019, Zhao2022} also follow the general framework of \cite{Negahban2012} for the same purpose. But their extension of this result to graph recovery does not permit zero threshold for edge detection. We attempt no such extension.
\vspace*{-0.15in}

\subsection{Our Contributions} 
In this paper, we analyze a group lasso penalized D-trace loss function approach for differential graph learning from MA data, extending the SA approach of \cite{Yuan2017, Jiang2018}. A two-block ADMM algorithm is presented to optimize the objective function. The two-block ADMM is guaranteed to be convergent unlike the three-block ADMM method used in \cite{Yuan2017}. Two different approaches to theoretical analysis of the proposed approach in high-dimensional settings are presented. Theorem 1 follows the approach(es) of \cite{Ravikumar2011, Kolar2014, Zhang2014, Yuan2017, Jiang2018} while Theorem 2 follows the general framework of \cite{Negahban2012}, not used in \cite{Yuan2017, Jiang2018}.  The general method of \cite{Ravikumar2011} requires an irrepresentability condition (see (\ref{eqn335})) which is also required in \cite{Yuan2017, Jiang2018} for SA graphs, but is not needed by the method of \cite{Negahban2012}, hence in our Theorem 2. Numerical results based on synthetic as well as real data are presented. 

Preliminary version of parts of this paper appear in a conference paper \cite{Tugnait23}. Theorem 2, proof of Theorem 1 and real data example do not appear in \cite{Tugnait23}.
\vspace*{-0.1in}

\subsection{Outline and Notation} \label{outnot}
The rest of the paper is organized as follows. A group lasso penalized D-trace loss function is presented in Sec.\ \ref{GM} for estimation of multi-attribute differential graph.  An ADMM algorithm is presented in Sec.\ \ref{SGL} to optimize the convex objective function. In Sec.\ \ref{TA} we analyze the properties of the estimator of the difference ${\bm \Delta} = {\bm \Omega}_y - {\bm \Omega}_x$. Theorem 1 follows the approach(es) of \cite{Ravikumar2011, Kolar2014, Zhang2014, Yuan2017, Jiang2018} while Theorem 2 follows the general framework of \cite{Negahban2012}. The general method of \cite{Ravikumar2011} requires an irrepresentability condition (see (\ref{eqn335})) which is not needed by the method of \cite{Negahban2012}. On the other hand, our Theorem 2 does not have a result like Theorem 1(ii), the oracle property, nor does it have a result as in Theorem 1(iv), support recovery. Numerical results based on synthetic as well as real data are presented in Sec.\ \ref{NE} to illustrate the proposed approach.  Proofs of Theorems 1 and 2 are given in Appendices \ref{append1} and \ref{append2}, respectively.

For a set $V$, $|V|$ or $\mbox{card}(V)$ denotes its cardinality. Given ${\bm A} \in \mathbb{R}^{p \times p}$, we use $\phi_{\min }({\bm A})$, $\phi_{\max }({\bm A})$, $|{\bm A}|$ and $\mbox{tr}({\bm A})$ to denote the minimum eigenvalue, maximum eigenvalue, determinant and  trace of ${\bm A}$, respectively. For ${\bm B} \in \mathbb{R}^{p \times q}$, we define  $\|{\bm B}\| = \sqrt{\phi_{\max }({\bm B}^\top  {\bm B})}$, $\|{\bm B}\|_F = \sqrt{\mbox{tr}({\bm B}^\top  {\bm B})}$, $\|{\bm B}\|_1 = \sum_{i,j} |B_{ij}|$, where $B_{ij}$ is the $(i,j)$-th element of ${\bm B}$ (also denoted by $[{\bm B}]_{ij}$), $\|{\bm B}\|_\infty = \max_{i,j} |B_{ij}|$ and $\|{\bm B}\|_{1,\infty} = \max_i \sum_j|B_{ij}|$. The symbols $\otimes$ and $\boxtimes$ denote Kronecker product and Tracy-Singh product \cite{Tracy1989}, respectively. In particular, given block partitioned matrices ${\bm A} =[{\bm A}_{ij}]$ and ${\bm B}=[{\bm B}_{k \ell}]$  with submatrices ${\bm A}_{ij}$ and ${\bm B}_{k \ell}$, Tracy-Singh product yields another block partitioned matrix ${\bm A} \boxtimes {\bm B} = [{\bm A}_{ij} \boxtimes {\bm B}]_{ij} = [[{\bm A}_{ij} \otimes {\bm B}_{k \ell}]_{k \ell} ]_{ij}$ \cite{Liu2008}. Given ${\bm A} =[{\bm A}_{ij}] \in \mathbb{R}^{mp \times mp}$ with ${\bm A}_{ij} \in \mathbb{R}^{m \times m}$, ${\rm vec}({\bm A}) \in \mathbb{R}^{m^2p^2}$ denotes the vectorization of ${\bm A}$ which stacks the columns of the matrix ${\bm A}$, and 
\begin{align*}
  & {\rm bvec}({\bm A}) =  [({\rm vec}({\bm A}_{11}))^\top \; ({\rm vec}({\bm A}_{21}))^\top \; \cdots 
      \; ({\rm vec}({\bm A}_{p1}))^\top \\
		& \quad ({\rm vec}({\bm A}_{12}))^\top \;  \cdots \; ({\rm vec}({\bm A}_{p2}))^\top \; \cdots \; ({\rm vec}({\bm A}_{pp}))^\top]^\top .
\end{align*} 

Let $S = {\cal E}_\Delta = \{ \{ k, \ell \} \,:\, \| {\bm \Delta}^{(k \ell)} \|_F \ne 0 \}$ where ${\bm \Delta} =[{\bm \Delta}^{(k \ell)}] \in \mathbb{R}^{mp \times mp}$ with ${\bm \Delta}^{(k \ell)} \in \mathbb{R}^{m \times m}$ denoting the $(k,l)$th $m \times m$ submatrix of ${\bm \Delta}$. Then ${\bm \Delta}_S$ denotes the submatrix of ${\bm \Delta}$ with block rows and columns indexed by $S$, i.e., ${\bm \Delta}_S =[{\bm \Delta}^{(k \ell)}]_{(k , \ell)  \in S}$. Suppose ${\bm \Gamma} = {\bm A} \boxtimes {\bm B}$ given block partitioned matrices ${\bm A} =[{\bm A}_{ij}]$ and ${\bm B}=[{\bm B}_{k \ell}]$. For any two subsets  $T_1$ and $T_2$ of  $V \times V$, ${\bm \Gamma}_{T_1,T_2}$ denotes the submatrix of ${\bm \Gamma}$ with block rows and columns indexed by $T_1$ and $T_2$, i.e., ${\bm \Gamma}_{T_1,T_2} = [{\bm A}_{j \ell} \otimes {\bm B}_{kq}]_{(j,k) \in T_1, (\ell,q) \in T_2}$. Following \cite{Kolar2014}, an operator $\bm{\mathcal C}( \cdot )$ is used in Sec.\ \ref{TA}. Consider ${\bm A} \in \mathbb{R}^{mp \times mp}$ with $(k,l)$th $m \times m$ submatrix ${\bm A}^{(k \ell)}$. Then $\bm{\mathcal C}( \cdot )$ operates on ${\bm A}$ as 
\begin{align*}
 \begin{bmatrix} {\bm A}^{(11)} & \cdots & {\bm A}^{(1p)} \\
			  \vdots     & \ddots     & \vdots \\
			{\bm A}^{(p1)} &  \cdots & {\bm A}^{(pp)} \end{bmatrix} \overset{\bm{\mathcal C}( \cdot )}{\xrightarrow{\hspace*{0.5cm}}}
			\begin{bmatrix} \|{\bm A}^{(11)}\|_F & \cdots & \|{\bm A}^{(1p)}\|_F \\
			  \vdots     & \ddots     & \vdots \\
			\|{\bm A}^{(p1)}\|_F &  \cdots & \|{\bm A}^{(pp)}\|_F \end{bmatrix}
\end{align*}
with $\bm{\mathcal C}( {\bm A}^{(k \ell)} ) = \|{\bm A}^{(k \ell)}\|_F$ and $\bm{\mathcal C}( {\bm A} ) \in \mathbb{R}^{p \times p}$. Now consider ${\bm A}_1  , {\bm A}_2 \in \mathbb{R}^{mp \times mp}$ with $(k,l)$th $m \times m$ submatrices ${\bm A}_1^{(k \ell)}$ and ${\bm A}_2^{(k \ell)}$, respectively, and Tracy-Singh product ${\bm A}_1 \boxtimes {\bm A}_2 \in \mathbb{R}^{(mp)^2 \times (mp)^2}$. Then $\bm{\mathcal C}( \cdot )$ operates on ${\bm A}_1 \boxtimes {\bm A}_2$ as $\bm{\mathcal C}({\bm A}_1 \boxtimes {\bm A}_2) \in \mathbb{R}^{p^2 \times p^2}$ with $\bm{\mathcal C}( {\bm A}_1^{(k_1 \ell_1)} \otimes {\bm A}_2^{(k_2 \ell_2)} ) = \|{\bm A}_1^{(k_1 \ell_1)} \otimes {\bm A}_2^{(k_2 \ell_2)} \|_F$ (=$\|{\bm A}_1^{(k_1 \ell_1)}  \|_F \, \|{\bm A}_2^{(k_2 \ell_2)} \|_F$). That is, each $m^2 \times m^2$ submatrix ${\bm A}_1^{(k_1 \ell_1)} \otimes {\bm A}_2^{(k_2 \ell_2)}$ of ${\bm A}_1 \boxtimes {\bm A}_2 $ is mapped into its Frobenius norm.
\vspace*{-0.1in}

\section{Group Lasso Penalized D-Trace Loss } \label{GM}
Let ${\bm x} = [ {\bm z}_{1x}^\top \; {\bm z}_{2x}^\top \; \cdots \; {\bm z}_{px}^\top ]^\top \in \mathbb{R}^{mp}$ where ${\bm z}_{ix} \in \mathbb{R}^m$, $i \in [p]$, are zero-mean, jointly Gaussian. Similarly, let ${\bm y} = [ {\bm z}_{1y}^\top \; {\bm z}_{2y}^\top \; \cdots \; {\bm z}_{py}^\top ]^\top \in \mathbb{R}^{mp}$ where ${\bm z}_{iy} \in \mathbb{R}^m$, $i \in [p]$, are zero-mean, jointly Gaussian. Given i.i.d.\ samples ${\bm x}(t)$, $t=1,2, \cdots , n_x$, of ${\bm x}$, and similarly given i.i.d.\ samples ${\bm y}(t)$, $t=1,2, \cdots , n_y$, of ${\bm y} \in \mathbb{R}^{mp}$, form the sample covariance estimates
\begin{equation}
  \hat{\bm \Sigma}_x = \frac{1}{n_x} \sum_{t=1}^{n_x} {\bm x}(t) {\bm x}^\top(t) \, , \;\;
	\hat{\bm \Sigma}_y = \frac{1}{n_y} \sum_{t=1}^{n_y} {\bm y}(t) {\bm y}^\top(t) \, .  \label{eqn10}
\end{equation} 
and denote their true values as ${\bm \Sigma}_x^\ast = {\bm \Omega}_x^{-\ast} (=({\bm \Omega}_x^\ast)^{-1})$ and ${\bm \Sigma}_y^\ast = {\bm \Omega}_y^{-\ast}$. Assume that $\{ {\bm x}(t) \}$ and $\{ {\bm y}(t) \}$ are mutually independent sequences. Assume ${\bm \Sigma}_x^\ast$ and ${\bm \Sigma}_y^\ast$ are positive definite. We wish to estimate ${\bm \Delta} = {\bm \Omega}_y^\ast - {\bm \Omega}_x^\ast$ and graph ${\cal G}_\Delta = \left( V, {\cal E}_\Delta \right)$, based on $\hat{\bm \Sigma}_x$ and $\hat{\bm \Sigma}_y$. Following the SA formulation of \cite{Yuan2017} (see also \cite[Sec.\ 2.1]{Jiang2018}), we will use a convex D-trace loss function given by
\begin{equation}
  L({\bm \Delta}, \hat{\bm \Sigma}_x , \hat{\bm \Sigma}_y) = \frac{1}{2} \mbox{tr} (\hat{\bm \Sigma}_x {\bm \Delta} \hat{\bm \Sigma}_y {\bm \Delta}^\top) 
	   - \mbox{tr} ({\bm \Delta}  (\hat{\bm \Sigma}_x-\hat{\bm \Sigma}_y)) \label{eqn15}
\end{equation}
where D-trace refers to difference-in-trace loss function, a term coined in \cite{Zhang2014} in the context of graphical model estimation. The  function $L({\bm \Delta}, {\bm \Sigma}_x^\ast , {\bm \Sigma}_y^\ast)$ is strictly convex in ${\bm \Delta}$ (its Hessian w.r.t.\ $\mbox{vec}({\bm \Delta})$ is ${\bm \Sigma}_y^\ast \otimes {\bm \Sigma}_x^\ast$), and has a unique minimum at ${\bm \Delta}^\ast = {\bm \Omega}_y^\ast - {\bm \Omega}_x^\ast$ \cite{Yuan2017, Jiang2018}. When we use sample covariances, we propose to estimate ${\bm \Delta}$ by minimizing the group-lasso penalized loss function
\begin{equation}
  L_\lambda({\bm \Delta}, \hat{\bm \Sigma}_x , \hat{\bm \Sigma}_y) = L({\bm \Delta}, \hat{\bm \Sigma}_x , \hat{\bm \Sigma}_y)
	   + \lambda \sum_{k, \ell=1}^p \| {\bm \Delta}^{(k \ell)} \|_F  \label{eqn20}
\end{equation}
where $\lambda > 0$ is a tuning parameter and $\| {\bm \Delta}^{(k \ell)} \|_F$ promotes blockwise sparsity in ${\bm \Delta}$ \cite{Yuan2006, Friedman2010a, Friedman2010b} where, if we partition ${\bm \Delta}$ into $m \times m$ submatrices, ${\bm \Delta}^{(k \ell)}$ denotes its $(k , \ell)$th submatrix, associated with edge $\{ k ,\ell\}$ of the differential graph ${\cal G}_\Delta = \left( V, {\cal E}_\Delta \right)$. 

For SA models ($m=1$), \cite{Jiang2018} has used the lasso-penalized loss function 
$L_{J}({\bm \Delta}) = L({\bm \Delta}, \hat{\bm \Sigma}_x , \hat{\bm \Sigma}_y)
	   + \lambda \sum_{k, \ell=1}^p \big| {\bm \Delta}_{k \ell} \big|$.  The cost $L_{J}({\bm \Delta})$ is optimized in \cite{Jiang2018} using a two-block ADMM approach which is known to be convergent. 
The resulting estimator $\hat{\bm \Delta}$ that minimizes the above cost is not necessarily symmetric. To obtain a symmetric estimator for SA models, \cite{Yuan2017} proposes the lasso-penalized loss function
$L_{Y}({\bm \Delta}) = \frac{1}{4} \mbox{tr}  (\hat{\bm \Sigma}_x {\bm \Delta} \hat{\bm \Sigma}_y {\bm \Delta}^\top 
	  + \hat{\bm \Sigma}_y {\bm \Delta} \hat{\bm \Sigma}_x {\bm \Delta}^\top)  - \mbox{tr} ({\bm \Delta}  (\hat{\bm \Sigma}_x-\hat{\bm \Sigma}_y)) + \lambda \sum_{k, \ell=1}^p \big| {\bm \Delta}_{k \ell} \big|$. 
In \cite{Yuan2017}, cost $L_{Y}({\bm \Delta})$ is optimized using a three-block ADMM method which is not necessarily convergent.

Suppose
\begin{equation}
  \hat{\bm \Delta} = \arg\min_{\bm \Delta} L_\lambda({\bm \Delta}, \hat{\bm \Sigma}_x , \hat{\bm \Sigma}_y) \, . \label{eqn25}
\end{equation}
Even though ${\bm \Delta}$ is symmetric, $\hat{\bm \Delta}$ is not. We can symmetrize it by setting $\hat{\bm \Delta}_{sym} = \frac{1}{2} ( \hat{\bm \Delta} + \hat{\bm \Delta}^\top )$, after obtaining $\hat{\bm \Delta}$. Then the differential graph edges are estimated as
\begin{equation}
  \hat{\cal E}_{\Delta} = \Big\{ \{k,\ell\} \, : \, \|\hat{\bm \Delta}_{sym}^{(k \ell)}\|_F > 0  \Big\} \, . \label{eqn26}
\end{equation}

\section{Optimization} \label{SGL}
The objective function $L_\lambda({\bm \Delta}, \hat{\bm \Sigma}_x , \hat{\bm \Sigma}_y)$, given by (\ref{eqn20}), is strictly convex. Several existing approaches such as an alternating direction method of multipliers (ADMM) \cite{Boyd2010} or proximal gradient descent (PGD) methods \cite{Beck2009}, can be followed to minimize (\ref{eqn20}). Note that \cite{Yuan2017, Jiang2018} use ADMM  while \cite{Tang2020} uses a proximal gradient method, all for SA graphs. It is stated in \cite[Sec.\ 2.2]{Yuan2017} that in their simulation example, ADMM approach yielded a slightly smaller value of the objective function compared to the PGD approach.  In \cite{Zhao2019, Zhao2022}, similar to \cite{Tang2020}, a proximal gradient method is used for an objective function similar to our (\ref{eqn20}). In this paper, motivated by \cite{Yuan2017}, we will develop an ADMM method. In a simulation example (Sec.\ \ref{NEsyn}) we compare our ADMM approach with ADMM and PGD approaches of  \cite{Jiang2018} and \cite{Tang2020, Zhao2022}, respectively.
\vspace*{-0.1in}

\subsection{ADMM Approach}
Similar to \cite{Jiang2018} (also \cite{Yuan2017}), we use an ADMM approach \cite{Boyd2010} with variable splitting. Using variable splitting, consider
\begin{align} 
 \min_{\bm{\Delta} , {\bm W} } & \Big\{  L({\bm \Delta}, \hat{\bm \Sigma}_x , \hat{\bm \Sigma}_y) 
   + \lambda \sum_{k, \ell=1}^p \| {\bm W}^{(k \ell)} \|_F  \Big\}  \label{eqn100}  \\
				& \quad	 \mbox{ subject to  }  \bm{\Delta} = {\bm W}  \, .       \nonumber
\end{align}
The scaled augmented Lagrangian for this problem is \cite{Boyd2010}
\begin{align} 
 L_\rho = & L({\bm \Delta}, \hat{\bm \Sigma}_x , \hat{\bm \Sigma}_y) 
   + \lambda \sum_{k, \ell=1}^p \| {\bm W}^{(k \ell)} \|_F  \nonumber \\ 
			& \quad		 + \frac{\rho}{2}  \| {\bm \Delta} - {\bm W} + {\bm U}\|^2_F   \label{eqn105}  
\end{align}
where ${\bm U}$ is the dual variable, and $\rho >0$ is the penalty parameter. Given the results $ \bm{\Delta}^{(i)}, {\bm W}^{(i)}, {\bm U}^{(i)}$ of the $i$th iteration, in the $(i+1)$st iteration, an ADMM algorithm executes the following three updates:
\begin{itemize}
\item[(a)] $\bm{\Delta}^{(i+1)} \leftarrow \arg \min_{\bm{\Delta}} \, L_a(\bm{\Delta}) ,\;\;
            L_a(\bm{\Delta}) := L({\bm \Delta}, \hat{\bm \Sigma}_x , \hat{\bm \Sigma}_y) + \frac{\rho}{2}  \| \bm{\Delta} 
						- {\bm W}^{(i)} + {\bm U}^{(i)}\|^2_F$.
\item[(b)] ${\bm W}^{(i+1)}  \leftarrow \arg \min _{ {\bm W} } L_b({\bm W}) , \;\;
          L_b({\bm W}) := \newline  \lambda \sum_{k, \ell=1}^p \| {\bm W}^{(k \ell)} \|_F 
					+ \frac{\rho}{2}  \| \bm{\Delta}^{(i+1)} - {\bm W} + {\bm U}^{(i)} \|^2_F$.
\item[(c)] ${\bm U}^{(i+1)} \leftarrow {\bm U}^{(i)}  +
   \left( \bm{\Delta}^{(i+1)} - {\bm W}^{(i+1)} \right)$.
\end{itemize}

\noindent {\bf Update (a)}: Differentiate $L_a(\bm{\Delta})$ w.r.t.\ ${\bm \Delta}$ to obtain
\begin{align} 
  {\bm 0} = & \frac{\partial L_a(\bm{\Delta})}{\partial {\bm \Delta}} =   
     \hat{\bm \Sigma}_x {\bm \Delta} \hat{\bm \Sigma}_y - (\hat{\bm \Sigma}_x-\hat{\bm \Sigma}_y) +
			\rho ( {\bm \Delta} - {\bm W} + {\bm U} ) \label{eqn115}  \\
	\Rightarrow & (\hat{\bm \Sigma}_y \otimes \hat{\bm \Sigma}_x + \rho {\bm I} )  {\rm vec}({\bm \Delta})
	      = {\rm vec}(\hat{\bm \Sigma}_x-\hat{\bm \Sigma}_y + \rho ({\bm W} - {\bm U}))  \label{eqn116}
\end{align}
Direct matrix inversion solution of (\ref{eqn116}) requires inversion of a $(mp)^2 \times (mp)^2$ matrix. A computationally cheaper solution is given in \cite{Yuan2017, Jiang2018}, as follows. Carry out eigendecomposition of $\hat{\bm \Sigma}_x$ and $\hat{\bm \Sigma}_y$ as $\hat{\bm \Sigma}_x = {\bm Q}_x {\bm D}_x {\bm Q}_x^\top$, ${\bm Q}_x {\bm Q}_x^\top = {\bm I}$ and $\hat{\bm \Sigma}_y = {\bm Q}_y {\bm D}_y {\bm Q}_y^\top$, ${\bm Q}_y {\bm Q}_y^\top = {\bm I}$, where ${\bm D}_x$ and ${\bm D}_y$ are diagonal matrices of the respective eigenvalues. Then $\hat{\bm \Delta}$ that minimizes $L_a(\bm{\Delta})$ is given by
\begin{align} 
  \hat{\bm \Delta} = & {\bm Q}_x \Big[ {\bm B} \circ [{\bm Q}_x^\top \big(\hat{\bm \Sigma}_x-\hat{\bm \Sigma}_y 
	         + \rho ( {\bm W} - {\bm U}) \big)
		   {\bm Q}_y ] \Big] {\bm Q}_y^\top  \label{eqn120}  
\end{align}
where the symbol $\circ$ denotes the Hadamard product and ${\bm B} \in \mathbb{R}^{mp \times mp}$ organizes the diagonal of $({\bm D}_y \otimes {\bm D}_x + \rho {\bm I})^{-1}$ in a matrix with ${\bm B}_{jk} = 1/([{\bm D}_x]_{jj} [{\bm D}_y]_{kk} + \rho)$. Note that the eigendecomposition of $\hat{\bm \Sigma}_x$ and $\hat{\bm \Sigma}_y$ has to be done only once. Thus
\begin{align} 
  {\bm \Delta}^{(i+1)} = & {\bm Q}_x \Big[ {\bm B} \circ [{\bm Q}_x^\top \big(\hat{\bm \Sigma}_x-\hat{\bm \Sigma}_y 
	     + \rho ({\bm W}^{(i)} - {\bm U}^{(i)}) \big) {\bm Q}_y ] \big] {\bm Q}_y^\top  \label{eqn125}  
\end{align}

\noindent {\bf Update (b)}: Here we have the group lasso solution \cite{Yuan2006, Friedman2010a, Friedman2010b}
\begin{align} 
  & \quad \quad ({\bm W}^{(k \ell)})^{(i+1)}  \nonumber \\ 
	= & \Big( 1 - \frac{(\lambda / \rho)}{  \| ({\bm \Delta}^{(i+1)}+ {\bm U}^{(i)})^{(k \ell)} \|_F } \Big)_+    
								  ({\bm \Delta}^{(i+1)}+ {\bm U}^{(i)})^{(k \ell)} \label{eqn130}  
\end{align}
where $(a)_+ = \max(0,a)$.

\begin{algorithm} 
\caption{ADMM Algorithm MA-ADMM}
\label{alg0}

\algorithmicrequire{\; Data $\{{\bm x}(t)\}_{t=1}^{n_x}$, ${\bm x} \in \mathbb{R}^{mp}$, and $\{{\bm y}(t)\}_{t=1}^{n_y}$, ${\bm y} \in \mathbb{R}^{mp}$, regularization and penalty parameters $\lambda$ and $\rho_0$, tolerances $\tau_{abs}$ and $\tau_{rel}$, variable penalty factor $\mu$, maximum number of iterations $i_{max}$.} \\
\algorithmicensure{\;\ estimated $\hat{\bm \Delta}_{sym}$ and $\hat{\cal E}_{\Delta}$.}

\begin{algorithmic}[1] 
\scriptsize  
\STATE Calculate sample covariances $\hat{\bm{\Sigma}}_x = \frac{1}{n_x} \sum_{t=1}^{n_x} {\bm x}(t) {\bm x}^\top(t)$ and $\hat{\bm{\Sigma}}_y = \frac{1}{n_y} \sum_{t=1}^{n_y} {\bm y}(t) {\bm y}^\top(t)$.
\STATE Initialize: ${\bm \Delta}^{(0)} = {\bm U}^{(0)} = {\bm W}^{(0)} = {\bm 0}$,  where ${\bm \Delta}, {\bm U}, {\bm W} \in \mathbb{R}^{(mp) \times (mp)}$, $\rho^{(0)} = \rho_0$.
\STATE Eigendecompose $\hat{\bm \Sigma}_x$ and $\hat{\bm \Sigma}_y$ as $\hat{\bm \Sigma}_x = {\bm Q}_x {\bm D}_x {\bm Q}_x^\top$ and $\hat{\bm \Sigma}_y = {\bm Q}_y {\bm D}_y {\bm Q}_y^\top$.
\STATE converged = FALSE, $i=0$
\WHILE{converged = FALSE $\;$ AND $\;$ $i \le i_{max}$,}
\STATE Construct ${\bm B} \in \mathbb{R}^{mp \times mp}$  with ${\bm B}_{jk} = 1/([{\bm D}_x]_{jj} [{\bm D}_y]_{kk} + \rho^{(i)}$.
\STATE Set ${\bm \Delta}^{(i+1)} =  {\bm Q}_x \Big[ {\bm B} \circ [{\bm Q}_x^\top \big(\hat{\bm \Sigma}_x-\hat{\bm \Sigma}_y + \rho ({\bm W}^{(i)} - {\bm U}^{(i)}) \big) {\bm Q}_y ] \big] {\bm Q}_y^\top \, $.
\STATE With $(a)_+ := \max(0,a)$, ${\bm A} = ({\bm \Delta}^{(i+1)}+ {\bm U}^{(i)})^{(k \ell)}$  and $k, \ell \in [p]$, update $m \times m$ subblocks of ${\bm W}$ as 
\begin{align*}  
  ({\bm W}^{(i+1)})^{(k \ell)}   
	= & \Big( 1 - \frac{(\lambda / \rho)}{  \| {\bm A} \|_F } \Big)_+    
								  {\bm A}^{(k \ell)}  \, .
\end{align*}
\STATE Dual update ${\bm U}^{(i+1)} = {\bm U}^{(i)} + \left( \bm{\Delta}^{(i+1)} - {\bm W}^{(i+1)} \right)$. 
\STATE Check convergence. Set tolerances
\begin{align*}
   \tau_{pri} = & mp \, \tau_{abs} + \tau_{rel} \, \max ( \| {\bm \Delta}^{(i+1)} \|_F , \| {\bm W}^{(i+1)} \|_F ) \\
  \tau_{dual} = & mp \, \tau_{abs} + \tau_{rel} \,  \| {\bm U}^{(i+1)} \|_F / \rho^{(i)} \, .
\end{align*}
Define $e_p = \| {\bm \Delta}^{(i+1)} - {\bm W}^{(i+1)} \|_F$ and $e_d = \rho^{(i)} \| {\bm W}^{(i+1)} - {\bm W}^{(i)} \|_F$.
If $( e_p \le \tau_{pri}) \; AND \; (e_d \le \tau_{dual})$, set converged = TRUE .
\STATE Update penalty parameter $\rho$ $\,$ : If $e_p > \mu e_d$, set $\rho^{(i+1)} = 2 \rho^{(i)}$, else if $e_d > \mu e_p$, set $\rho^{(i+1)} = \rho^{(i)}/2$, otherwise $\rho^{(i+1)} =  \rho^{(i)}$.
We also need to set ${\bm U}^{(i+1)} = {\bm U}^{(i+1)}/2$ for $e_p > \mu e_d$ and ${\bm U}^{(i+1)} = 2 {\bm U}^{(i+1)}$ for $e_d > \mu e_p$.
\STATE $i \leftarrow i+1$
\ENDWHILE
\STATE Set $\hat{\bm \Delta}_{sym} = \frac{1}{2} ({\bm W} + {\bm W}^\top)$. If $\|\hat{\bm \Delta}_{sym}^{(jk)}\|_F > 0$, assign edge $\{ j,k\} \in \hat{\cal E}_{\Delta}$, else $\{ j,k\} \not\in \hat{\cal E}_{\Delta}$. 
\end{algorithmic}
\end{algorithm}

\begin{algorithm} 
\caption{ADMM Algorithm SA-ADMM}
\label{alg1}

\algorithmicrequire{\; As in Algorithm \ref{alg0}. }\\
\algorithmicensure{\;\ Estimated $\hat{\bm \Delta}_{sym}$ and $\hat{\cal E}_{\Delta}$.}

\begin{algorithmic}[1] 
\scriptsize
\STATE Follow lines 1-7 of Algorithm \ref{alg0}
\STATE With $(a)_+ := \max(0,a)$,  ${\bm A} = ({\bm \Delta}^{(i+1)}+ {\bm U}^{(i)})^{(k \ell)}$ and $k, \ell \in [mp]$, update $[{\bm W}]_{k \ell} \in \mathbb{R}$ as 
\begin{align*}  
  [{\bm W}^{(i+1)}]_{k \ell}   
	= & \Big( 1 - \frac{(\lambda / \rho)}{ \big| [{\bm A}]_{k \ell} \big| } \Big)_+    
								  [{\bm A}]_{k \ell}  \, .
\end{align*}
\STATE Follow lines 9-14 of Algorithm \ref{alg0}
\end{algorithmic}
\end{algorithm}

A pseudocode for the ADMM algorithm, MA-ADMM, used in this paper is given in Algorithm \ref{alg0} where we use the stopping (convergence) criterion following \cite[Sec.\ 3.3.1]{Boyd2010} and varying penalty parameter $\rho$ following \cite[Sec.\ 3.4.1]{Boyd2010}. The stopping criterion is based on primal and dual residuals being small where, in our case, at $(i+1)$st iteration, the primal residual is given by $\bm{\Delta}^{(i+1)} - {\bm W}^{(i+1)}$ and the dual residual by $\rho^{(i)} ({\bm W}^{(i+1)} - {\bm W}^{(i)})$. Convergence criterion is met when the norms of these residuals are below primary and dual tolerances $\tau_{pri}$ and $\tau_{dual}$, respectively; see line 10 of Algorithm \ref{alg0}. In turn, $\tau_{pri}$ and $\tau_{dual}$ are chosen using an absolute and relative criterion as in line 10 of Algorithm \ref{alg0} where $\tau_{abs}$ and $\tau_{rel}$ are user chosen absolute and relative tolerances, respectively.  Line 10 of Algorithm \ref{alg0} follows typical choices given in \cite[Sec.\ 3.4.1]{Boyd2010}. For all numerical results presented later, we used $\rho_0 =2$, $\mu =10$, and $\tau_{abs}=\tau_{rel} =10^{-4}$.

We will compare our approach with three other approaches in Sec.\ \ref{NEsyn}. One of them is the single attribute (SA) based ADMM approach (see \cite{Yuan2017, Jiang2018}. A pseudocode of our implementation of this approach, SA-ADMM, is in Algorithm \ref{alg1} which differs from in Algorithm \ref{alg0} only in line 8 where we replace group lasso with elementwise lasso.

\subsection{Proximal Gradient Descent Approach} \label{PGD}  
It is a first-order method that is based on objective function values and gradient evaluations. A pseudocode of the PGD method of \cite{Zhao2022}, MA-proximal, is in Algorithm \ref{alg2}, and that of \cite{Tang2020}, SA-proximal, is in Algorithm \ref{alg3}. Algorithm \ref{alg3} differs from in Algorithm \ref{alg2} only in line 8 where we replace group lasso with element-wise lasso. For all numerical results presented later, we used $\epsilon =10^{-3}$ in line 7.

\begin{algorithm} 
\caption{PGD Algorithm MA-PGD}
\label{alg2}

\algorithmicrequire{\; Data $\{{\bm x}(t)\}_{t=1}^{n_x}$, ${\bm x} \in \mathbb{R}^{mp}$, and $\{{\bm y}(t)\}_{t=1}^{n_y}$, ${\bm y} \in \mathbb{R}^{mp}$, tolerance $\epsilon$, maximum number of iterations $i_{max}$} \\
\algorithmicensure{\;\ Estimated $\hat{\bm \Delta}_{sym}$ and $\hat{\cal E}_{\Delta}$.}

\begin{algorithmic}[1] 
\scriptsize
\STATE Calculate sample covariances $\hat{\bm{\Sigma}}_x = \frac{1}{n_x} \sum_{t=1}^{n_x} {\bm x}(t) {\bm x}^\top(t)$ and $\hat{\bm{\Sigma}}_y = \frac{1}{n_y} \sum_{t=1}^{n_y} {\bm y}(t) {\bm y}^\top(t)$.
\STATE Set $\eta = 1/\big(\phi_{max}(\hat{\bm \Sigma}_x) \, \phi_{max}(\hat{\bm \Sigma}_x) \big)$. Initialize: ${\bm \Delta}^{(0)} =  {\bm 0}$.
\STATE converged = FALSE, $i=0$
\WHILE{converged = FALSE $\;$ AND $\;$ $i \le i_{max}$,}
\STATE Set ${\bm A} =  {\bm \Delta}^{(i)} - \eta \Big( \hat{\bm \Sigma}_x {\bm \Delta}^{(i)} \hat{\bm \Sigma}_y 
 - (\hat{\bm \Sigma}_x-\hat{\bm \Sigma}_y) \Big)$.
\STATE For $k, \ell \in [p]$, update $m \times m$ subblocks as
\begin{align*}  
  ({\bm \Delta}^{(i+1)})^{(k \ell)}   
	= & \Big( 1 - \frac{\lambda \eta}{  \| {\bm A} \|_F } \Big)_+    
								  {\bm A}^{(k \ell)}  \, .
\end{align*}
\STATE  If $\frac{L_\lambda({\bm \Delta}^{(i+1)}, \hat{\bm \Sigma}_x , \hat{\bm \Sigma}_y)-L_\lambda({\bm \Delta}^{(i)}, \hat{\bm \Sigma}_x , \hat{\bm \Sigma}_y)}{L_\lambda({\bm \Delta}^{(i)}, \hat{\bm \Sigma}_x , \hat{\bm \Sigma}_y)} \le \epsilon$, set converged = TRUE .
\STATE $i \leftarrow i+1$
\ENDWHILE
\STATE Set $\hat{\bm \Delta}_{sym} = \frac{1}{2} ({\bm \Delta} + {\bm \Delta}^\top)$. If $\|\hat{\bm \Delta}_{sym}^{(jk)}\|_F > 0$, assign edge $\{ j,k\} \in \hat{\cal E}_{\Delta}$, else $\{ j,k\} \not\in \hat{\cal E}_{\Delta}$.
\end{algorithmic}
\end{algorithm} 
\vspace*{-0.1in}

\begin{algorithm} 
\caption{PGD Algorithm SA-PGD}
\label{alg3}

\algorithmicrequire{\; As in Algorithm \ref{alg2} }\\
\algorithmicensure{\;\ Estimated $\hat{\bm \Delta}$ and $\hat{\cal E}_{\Delta}$}

\begin{algorithmic}[1] 
\scriptsize
\STATE Follow lines 1-5 of Algorithm \ref{alg2}
\STATE For $k, \ell \in [mp]$, update 
\begin{align*}  
  [{\bm \Delta}^{(i+1)}]_{k \ell}   
	= & \Big( 1 - \frac{\lambda \eta}{ \big| [{\bm A}]_{k \ell} \big| } \Big)_+    
								  [{\bm A}]_{k \ell}  \, .
\end{align*}
\STATE Follow lines 7-10 of Algorithm \ref{alg2}
\end{algorithmic}
\end{algorithm} 
\vspace*{-0.1in}

\subsection{Computational Complexity}  \label{CC} 
The computational complexity of ADMM and PGD methods has been discussed in \cite{Tang2020} for SA differential graphs, and it is of the same order for MA graphs, because the difference lies only in lasso versus group lasso, i.e., element-wise soft-thresholding versus group-wise soft-thresholding. Noting that we have $mp \times mp$ precision matrices, by \cite{Tang2020}, the computational complexity of the ADMM approaches (our proposed and that of \cite{Yuan2017, Jiang2018}) is ${\cal O}((mp)^3)$ while that of the PGD methods of \cite{Tang2020, Zhao2022} is  either ${\cal O}((mp)^3)$ when as implemented in Algorithms \ref{alg2} and \ref{alg3}, or ${\cal O}((n_x+n_y)(mp)^2)$ when an alternative implementation of the cost gradient in line 5 of Algorithms \ref{alg2} and \ref{alg3} is used (see \cite[Sec.\ 2.2]{Tang2020}). For $n_x+n_y \ge mp$, there is no advantage to this alternative approach.
\vspace*{-0.1in}

\subsection{Convergence of ADMM} \label{conv} The objective function  (\ref{eqn20}), is strictly convex. It is also closed, proper and lower semi-continuous. Hence, for any fixed $\rho > 0$, the (two-block) ADMM algorithm is guaranteed to converge \cite[Sec.\ 3.2]{Boyd2010}, in the sense that we have primal residual convergence to 0, dual residual convergence to 0, and objective function convergence to the optimal value. For varying $\rho$, the convergence of ADMM has not been proven, but if we additionally impose $\rho^{(i)} = \rho^{(i_0)} > 0$ for $ i \ge i_0$ for some $i_0$, we have convergence \cite[Sec.\ 3.4.1]{Boyd2010}.
\vspace*{-0.1in}

\subsection{Model Selection} \label{modelsel}  Following the lasso penalty work of \cite{Yuan2017} (who invokes \cite{Zhao2014}), we use the following criterion for selection of group lasso penalty:
\begin{align} 
  BIC(\lambda) = & (n_x+n_y) \, \| \hat{\bm \Sigma}_x \hat{\bm \Delta} \hat{\bm \Sigma}_y - (\hat{\bm \Sigma}_x
	   - \hat{\bm \Sigma}_y ) \|_F \nonumber \\
		& \quad + \ln (n_x+n_y) \, | \hat{\bm \Delta} |_0
  \label{eqn180}  
\end{align}
where $| {\bm A} |_0$ denotes number of nonzero elements in ${\bm A}$ and $\hat{\bm \Delta}$ obeys (\ref{eqn25}). Choose $\lambda$ to minimize $BIC(\lambda)$. Following \cite{Yuan2017} we term it BIC (Bayesian information criterion) even though the cost function used is not negative log-likelihood although $\ln (n_x+n_y) \, | \hat{\bm \Delta} |_0$ penalizes over-parametrization as in BIC. It is based on the fact that true ${\bm \Delta}^\ast$ satisfies ${\bm \Sigma}_x^\ast {\bm \Delta}^\ast {\bm \Sigma}_y^\ast - ({\bm \Sigma}_x^\ast- {\bm \Sigma}_y^\ast ) = {\bm 0}$. Since (\ref{eqn180}) is not scale invariant, we scale both $\hat{\bm \Sigma}_x$ and $\hat{\bm \Sigma}_y$ (and $\hat{\bm \Delta}$ commensurately) by $\bar{\bm \Sigma}^{-1}$ where $\bar{\bm \Sigma} = \mbox{diag}\{\hat{\bm \Sigma}_x\}$ is a diagonal matrix of diagonal elements of $\hat{\bm \Sigma}_x$.
 
In our simulations we search over $\lambda \in [\lambda_{\ell} , \lambda_{u}]$,  where $\lambda_{\ell}$ and $\lambda_u$ are selected via a heuristic as in \cite{Tugnait21a}. Find the smallest $\lambda$, labeled $\lambda_{sm}$ for which we get a no-edge model; then we set $\lambda_{u}= \lambda_{sm}/2$ and $\lambda_{\ell} = \lambda_{u}/10$.

\section{Theoretical Analysis} \label{TA}  
Here we analyze the properties of $\hat{\bm \Delta}$. Theorem 1 follows the approach(es) of \cite{Ravikumar2011, Kolar2014, Zhang2014, Yuan2017, Jiang2018} while Theorem 2 follows the general framework of \cite{Negahban2012}. The general method of \cite{Ravikumar2011} used in \cite{Kolar2014, Zhang2014, Yuan2017, Jiang2018} requires an irrepresentability condition (see (\ref{eqn335})) which is not needed by the method of \cite{Negahban2012}. On the other hand, our Theorem 2 does not have a result like Theorem 1(ii), the oracle property, or support recovery Theorem 1(iv).

First some notation. Define the true differential edgeset
\begin{align} 
  S = & {\cal E}_{\Delta^\ast} = \{ \{ k, \ell \} \,:\, \| {\bm \Delta}^{\ast (k \ell)} \|_F \ne 0 \} \, , \quad
	    s = |S|  \, .
  \label{eqn200}  
\end{align}
Define 
\begin{equation}
  {\bm \Gamma}^\ast = {\bm \Sigma}_y^\ast \boxtimes {\bm \Sigma}_x^\ast \, , \quad
	\hat{\bm \Gamma} = \hat{\bm \Sigma}_y \boxtimes \hat{\bm \Sigma}_x \, . \label{eqn225}
\end{equation}
Also, recall the  operator $\bm{\mathcal C}( \cdot )$ defined in Sec.\ \ref{outnot}.  
In the rest of this section, we allow $p$, $s$ and $\lambda$ to be a functions of sample size $n$, denoted as $p_n$, $s_n$ and $\lambda_n$, respectively.  Define
\begin{align}
  M & = \max \{ \| \bm{\mathcal C}({\bm \Sigma}_x^\ast) \|_\infty \, , 
	     \| \bm{\mathcal C}({\bm \Sigma}_y^\ast) \|_\infty \} \, , \label{eqn320} \\
	M_\Sigma & = \max \{ \| \bm{\mathcal C}({\bm \Sigma}_x^\ast) \|_{1,\infty} \, , 
	           \| \bm{\mathcal C}({\bm \Sigma}_y^\ast) \|_{1,\infty} \} \, , \label{eqn325} \\
\kappa_\Gamma & = \| \bm{\mathcal C}(\Gamma_{S,S}^\ast)^{-1} \|_{1,\infty} \, ,\label{eqn330} \\
	\alpha & = 1- \max_{e \in S^c} \| \bm{\mathcal C}({\bm \Gamma}_{e,S}^\ast ({\bm \Gamma}_{S,S}^\ast)^{-1}) \|_{1}  \, ,
	     \label{eqn335} \\
	\bar{\sigma}_{xy} &  = \max \{ \max_i [{\bm \Sigma}_{x}^\ast]_{ii}, \, \max_i [{\bm \Sigma}_{y}^\ast]_{ii} \} 
	     \label{eqn336}  \\
	{C}_0 & = 40 \, m \, \bar{\sigma}_{xy}  \sqrt{2 \big(\tau + \ln(4m^2)/\ln(p_n) \big)}
	     \label{eqn337} 
\end{align}
where $S$ and ${\bm \Gamma}^\ast$ have been defined in (\ref{eqn200}) and (\ref{eqn225}). In (\ref{eqn335}), we require $0 < \alpha < 1$, and the expression 
\[
   \max_{e \in S^c} \| \bm{\mathcal C}({\bm \Gamma}_{e,S}^\ast ({\bm \Gamma}_{S,S}^\ast)^{-1}) \|_{1} \le 1-\alpha
\]
for some $\alpha \in (0,1)$ is called the {\it irrepresentability condition}. Similar conditions are also used in \cite{Ravikumar2011, Kolar2014, Zhang2014, Yuan2017, Jiang2018}.

Let $\hat{\bm \Delta}$ be as in (\ref{eqn25}). \\
{\bf Theorem 1}. For the system model of Sec.\ \ref{GM}, under (\ref{eqn200}) and the irrepresentability condition (\ref{eqn335}) for some $\alpha \in (0,1)$, if
\begin{align}  
 &  \lambda_n =  \max \Big\{ \frac{8}{\alpha} \, , \frac{3}{\alpha \bar{C}_\alpha} \, s_n \kappa_\Gamma M  C_{M\kappa} \Big\}
		     {C}_0 \sqrt{\frac{\ln(p_n)}{n}}   \label{eqn350} \\ 
& n = \min(n_x,n_y) >   	\, \max \Big\{\frac{1}{\min\{M^2,1\}} \, , 
              81 M^2 s_n^2 \kappa_\Gamma^2, \nonumber \\
  & \quad\quad \quad\quad \frac{9 s_n^2}{(\alpha \bar{C}_\alpha)^2}  (\kappa_\Gamma M C_{M\kappa})^2 \Big\} 
	  {C}_0^2 \ln(p_n)
		     \label{eqn352}
\end{align}
where $\bar{C}_\alpha = \frac{1-\alpha}{2(2M+1)-2 \alpha M }$ and $C_{M\kappa} = 1.5 \big(1+ \kappa_\Gamma \min\{s_n M^2,M_\Sigma^2\} \big)$, then with probability $> 1- 2/p_n^{\tau -2}$, for any $\tau >2$, we have
\begin{itemize}
\item[(i)] $\| \bm{\mathcal C}(\hat{\bm \Delta} - {\bm \Delta}^\ast) \|_\infty \le (C_{b1} + C_{b2}) {C}_0 \sqrt{\frac{\ln(p_n)}{n}}$ 
\begin{align*}
 \mbox{where } \;   C_{b1} = & 3 \kappa_\Gamma \, \max \big\{ \frac{8}{\alpha} \, , 
	 \frac{3}{\alpha \bar{C}_\alpha} \, s_n \kappa_\Gamma M C_{M\kappa} \big\} \\
	C_{b2} = & 9 s_n \kappa_\Gamma^2 M^2 \, . 
\end{align*}
\item[(ii)] $\hat{\bm \Delta}_{S^c} = {\bm 0}$.
\item[(iii)] $\| \bm{\mathcal C}(\hat{\bm \Delta} - {\bm \Delta}^\ast) \|_F \le \sqrt{s_n} \, \| \bm{\mathcal C}(\hat{\bm \Delta} - {\bm \Delta}^\ast) \|_\infty$ .
\item[(iv)] Additionally, if $\min_{(k,\ell) \in S} \| ({\bm \Delta}^\ast)^{(k \ell)} \|_F \ge  \newline 2 (C_{b1} + C_{b2}) {C}_0 \sqrt{\frac{\ln(p_n)}{n}}$, then $P({\cal G}_{\hat{\Delta}} = {\cal G}_{{\Delta}^\ast}) > 1- 2/p_n^{\tau -2}$ (support recovery) $\quad \bullet$
\end{itemize}
The proof of Theorem 1 is given in Appendix \ref{append1}.

Now we present Theorem 2 that follows the general framework of \cite{Negahban2012}. Let $\hat{\bm \Delta}$ be as in (\ref{eqn25}). \\
{\bf Theorem 2}. For the system model of Sec.\ \ref{GM}, under (\ref{eqn200}), if
\begin{align}  
 &  \lambda_n \ge  (4+6 M s_n C_1 )
		     {C}_0 \sqrt{\frac{\ln(p_n)}{n}}   \label{eqn450} \\ 
& n = \min(n_x,n_y) >   	\, \max \Big\{\frac{1}{M^2} \, , 
              \big( \frac{96 M s_n}{\phi^\ast_{min}} \big)^2 \Big\} 
	  {C}_0^2 \ln(p_n)
		     \label{eqn452}
\end{align}
where $C_1 = \max_{\{k, \ell \} \in V \times V} \| ({\bm \Delta}^\ast)^{(k \ell)} \|_F$ and $\phi^\ast_{min} = \phi_{min}({\bm \Sigma}_x^\ast) \phi_{min}({\bm \Sigma}_y^\ast)$, then with probability $> 1- 2/p_n^{\tau -2}$, for any $\tau >2$, we have
\begin{align}
  & \| \hat{\bm \Delta} - {\bm \Delta}^\ast \|_F \le \frac{12 \sqrt{s_n}}{\phi^\ast_{min}}
	  (4+6 M s_n C_1 ) {C}_0 \sqrt{\frac{\ln(p_n)}{n}}   \quad \bullet  \label{eqn454}
\end{align}

The proof of Theorem 2 is given in Appendix \ref{append2}. Note that $\| \bm{\mathcal C}(\hat{\bm \Delta} - {\bm \Delta}^\ast) \|_F = \| \hat{\bm \Delta} - {\bm \Delta}^\ast \|_F$ when comparing Theorems 1 and 2. \\
{{\bf Remark 1}: {\it Convergence Rate}}. If $M$, $M_\Sigma$ and $\kappa_\Gamma$ stay bounded with increasing sample size $n$, we have $\| \bm{\mathcal C}(\hat{\bm \Delta} - {\bm \Delta}^\ast) \|_F = {\cal O}_P (s_n^{1.5} \sqrt{ \ln(p_n)/n})$. Therefore, for $\| \bm{\mathcal C}(\hat{\bm \Delta} - {\bm \Delta}^\ast) \|_F \rightarrow 0$ as $n \rightarrow \infty$, we must have $s_n^{1.5} \sqrt{ \ln(p_n)/n} \rightarrow 0$. The SA results in \cite{Yuan2017} need $s_n^{3.5} \sqrt{ \ln(p_n)/n} \rightarrow 0$ when we take into account the dependence of various constants on $s_n$ in \cite{Yuan2017}. Notice that $M_\Sigma$ constraints covariances ${\bm \Sigma}_x^\ast$ and ${\bm \Sigma}_y^\ast$ which can be dense even if ${\bm \Omega}_x^\ast$ and ${\bm \Omega}_y^\ast$ are sparse (they need not be sparse for differential estimation), making them possibly unbounded with increasing sample size $n$. In this case we use $\min\{s_n M^2,M_\Sigma^2\}= s_n M^2$ in $C_{M\kappa}$ and $C_{b1}$, with $M$ bounded, leading to $\| \bm{\mathcal C}(\hat{\bm \Delta} - {\bm \Delta}^\ast) \|_F = {\cal O}_P (s_n^{2.5} \sqrt{ \ln(p_n)/n})$. On the other hand, in Theorem 2, we always have $\| \bm{\mathcal C}(\hat{\bm \Delta} - {\bm \Delta}^\ast) \|_F = {\cal O}_P (s_n^{1.5} \sqrt{ \ln(p_n)/n})$. $\;\; \Box$

{\bf Remark 2}: Our results assume Gaussian data. Theorems 1 and 2 continue to hold for more general sub-Gaussian distributions (Gaussian distribution is a sub-Gaussian distribution) except that the zeros in the precision matrix (see (\ref{neweq24})), or in the difference of two precision matrices, no longer signify conditional independence (or change in conditional independence) of the random vectors associated with the respective nodes; they only imply zero partial correlation. We state Lemma 1 in Appendix \ref{append1} for sub-Gaussian distributions, following \cite[Lemma 1]{Ravikumar2011}. Lemma 2 is then specialized to Gaussian distributions by setting the sub-Gaussian parameter $\sigma_{sg} =1$. If $\sigma_{sg} \ne 1$, then the only changes required in Theorems 1 and 2 is a scaling of $C_0$ in (\ref{eqn337}) where one needs to replace the factor of 40 with $8 (1 + 4 \sigma_{sg}^2)$. 
$\;\; \Box$

{\bf Remark 3}: Theorems 1 and 2 assume a constant number of attributes $m$, with only $p$, $s$ and $\lambda$ allowed to be functions of sample size $n$, and our Remark 1 reflects this fact. In terms of $m$, the bounds in Theorem 1(i) and Theorem 2 are ${\cal O}(m^3)$, which follows from $M = {\cal O}(m)$ and $C_{M\kappa} = {\cal O}(m^2)$ in Theorem 1, and  $M = {\cal O}(m)$, $C_1 = {\cal O}(m)$ and $C_0  = {\cal O}(m)$ in Theorem 2.  
Therefore, for large $m$, one would need much higher number of samples $n$, and it is not clear how to circumvent this bottleneck.  A different class of models based on matrix-valued graphical modeling \cite{Leng2012, Tsiligkaridis2013, Lyu2020} is a potential solution. In a matrix graph set-up, one has matrix-valued observations ${\bm Z}$, which in our context would require attributes (components of ${\bm z}_i$) to be arranged along rows and nodes $i$ along columns, with the covariance of $\mbox{vec}({\bm Z})$ having a Kronecker-product structure. This structure drastically reduces the number of unknowns from ${\cal O}((mp)^2)$ to ${\cal O}(m^2+p^2)$. Prior reported work (\cite{Leng2012, Tsiligkaridis2013, Lyu2020}) is on matrix graph estimation, with no reported work on differential matrix graphs.  
$\;\; \Box$
\vspace*{-0.1in}

\section{Numerical Examples} \label{NE}
We now present numerical results for both synthetic and real data to illustrate the proposed approach. In synthetic data examples the ground truth is known and this allows for assessment of the efficacy of various approaches. In real data examples where the ground truth is unknown, our goal is visualization and exploration of the differential conditional dependency structures underlying the data.
\vspace*{-0.1in}

\subsection{Synthetic Data: Erd\"{o}s-R\`{e}nyi and Barab\'{a}si-Albert Graphs} \label{NEsyn}
We consider two types of graphs: Erd\"{o}s-R\`{e}nyi (ER) graph and Barab\'{a}si-Albert (BA) graph \cite{Barabasi1999, Lu2014}. The BA graphs are an example of scale-free graphs with power law degree distribution \cite{Barabasi1999}. 
In the ER graph, $p=100$ nodes are connected to each other with probability $p_{er} =0.5$ and there are $m=3$ attributes per node whereas in the BA graph, we used $p=100$ and mean degree of 2 to generate a BA graph using the procedure given in \cite{Lu2014} (MATLAB function BAmodel.m from \url{https://github.com/ShanLu1984/Scale-Free-Network-Generation-and-Comparison}). In the upper triangular $\bm{\Omega}_x$, we set $[\bm{\Omega}_x^{(jk)}]_{st} = 0.5^{|s-t|}$ for $j=k=1, \cdots, p$, $s,t=1, \cdots , m$. For $j \ne k$, if the two nodes are not connected in the graph (ER or BA), we have  $\bm{\Omega}^{(jk)} = {\bm 0}$, and if nodes $j$ and $k$ are connected, then $[\bm{\Omega}^{(jk)}]_{st}$ is uniformly distributed over $[-0.4,-0.1] \cup [0.1,0.4]$. Then add lower triangular elements to make $\bm{\Omega}_x$ a symmetric matrix. To generate $\bm{\Omega}_y$, we follow \cite{Yuan2017} and first generate a differential graph with ${\bm \Delta} \in \mathbb{R}^{(mp) \times (mp)}$ as an ER graph (regardless of whether $\bm{\Omega}_x$ is based on ER or BA model), with connection probability $p_{er} =0.05$ (sparse): if nodes $j$ and $k$ are connected in the $\bm{\Omega}_x$ model, then each of $m^2$ elements of $\bm{\Delta}^{(jk)}$ is independently set to $\pm 0.9$ with equal probabilities. Then $\bm{\Omega}_y = \bm{\Omega}_x + {\bm \Delta}$. Finally add $\gamma {\bm I}$ to  $\bm{\Omega}_y$ and to $ \bm{\Omega}_x$ and pick $\gamma$ so that $\bm{\Omega}_y$ and $\bm{\Omega}_x$ are both positive definite.  With $\bm{\Phi}_x \bm{\Phi}_x^\top =\bm{\Omega}_x^{-1}$, we generate ${\bm x} = \bm{\Phi} {\bm w}$ with ${\bm w} \in \mathbb{R}^{mp}$ as zero-mean Gaussian, with identity covariance, and similarly for ${\bm y}$.
We generate $n=n_x=n_y$ i.i.d.\ observations for ${\bm x}$ and ${\bm y}$, with $m=3$, $p =100$, $n \in \{100, 200, 300, 400, 800, 1200, 1600\}$.
\vspace*{-0.07in}
\begin{figure}[htb]
  \centering
  \includegraphics[width=0.7\linewidth]{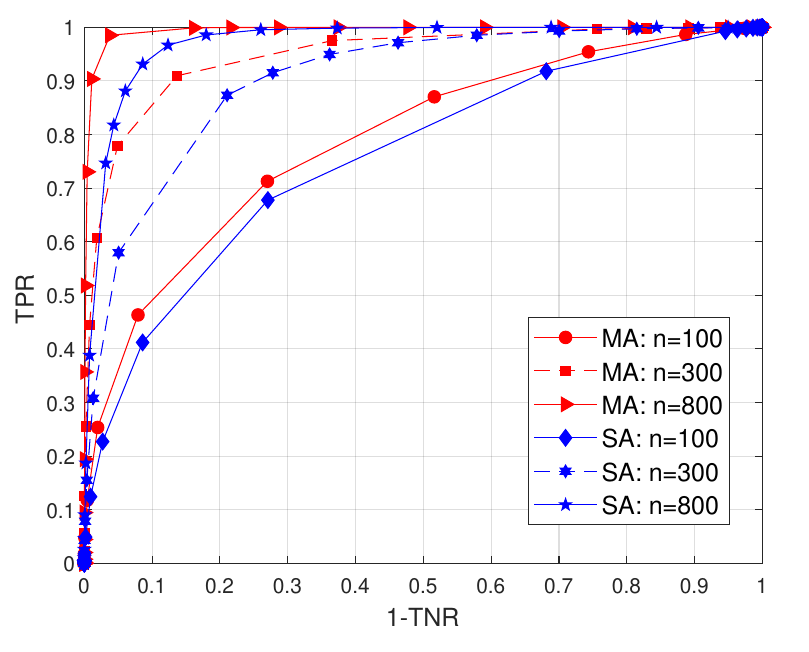} 
	\vspace*{-0.05in}
  \caption{ROC curves for ER graph based on ADMM approaches. TPR=true positive rate, TNR=true negative rate}
  \label{fig1}
\end{figure}
\vspace*{-0.17in}
\begin{figure}[htb]
  \centering
  \includegraphics[width=0.7\linewidth]{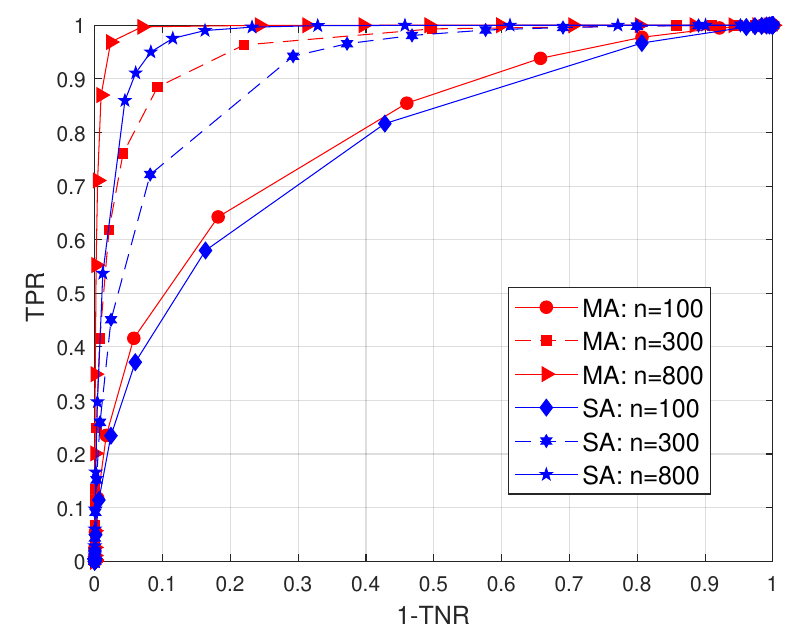} 
	\vspace*{-0.05in}
  \caption{ROC curves for BA graph based on ADMM approaches. TPR=true positive rate, TNR=true negative rate}
  \label{fig2}
\end{figure}  

\begin{table*}
\begin{center}
\begin{tabular}{ccccc}   \hline\hline
 Approach  & $F_1$ score ($\pm \sigma$)  & timing (s) ($\pm \sigma$) & TPR ($\pm \sigma$) & 1-TNR ($\pm \sigma$) \\  \hline
\multicolumn{5}{c}{ $n$ = 300 } \\ \hline
MA-ADMM & 0.6152 $\pm$0.0705&  2.5044 $\pm$0.2939& 0.6067 $\pm$0.1230&  0.0184 $\pm$0.0080 \\ 
MA-proximal & 0.6686 $\pm$0.0639&  6.6931 $\pm$0.3743 & 0.6845 $\pm$0.1253&  0.0184 $\pm$0.0085 \\  \hline
SA-ADMM& 0.4549 $\pm$0.0332&  0.1739 $\pm$0.0173& 0.5795 $\pm$0.1132&  0.0506 $\pm$0.0186 \\
SA-proximal & 0.4772 $\pm$0.0328&  0.3517 $\pm$0.0195 & 0.6263 $\pm$0.1157&  0.0524 $\pm$0.0193\\ \hline
\multicolumn{5}{c}{ $n$ = 800 } \\ \hline
MA-ADMM & 0.8537 $\pm$0.0491&  2.1911 $\pm$0.0639& 0.9037 $\pm$0.0703&  0.0111 $\pm$0.0041 \\ 
MA-proximal & 0.8898 $\pm$0.0408&  4.8277 $\pm$0.2647 & 0.9526 $\pm$0.0537&  0.0009 $\pm$0.0041 \\  \hline
SA-ADMM& 0.6336 $\pm$0.0055&  0.1510 $\pm$0.0081& 0.7468 $\pm$0.1017&  0.0316 $\pm$0.0090 \\
SA-proximal & 0.6612 $\pm$0.0401&  0.2533 $\pm$0.0143 & 0.7917 $\pm$0.0931&  0.0314 $\pm$0.0090\\ \hline
\multicolumn{5}{c}{ $n$ = 3000 } \\ \hline
MA-ADMM & 0.9795 $\pm$0.0152&  2.1436 $\pm$0.0624& 0.9928 $\pm$0.0018&  0.0012 $\pm$0.0041 \\ 
MA-proximal & 0.9914 $\pm$0.0106&  3.8312 $\pm$0.3948 & 0.9964 $\pm$0.0121&  0.0007 $\pm$0.0007 \\  \hline
\multicolumn{5}{c}{ $n$ = 6000 } \\ \hline
MA-ADMM & 0.9914 $\pm$0.0087&  1.9459 $\pm$0.0544& 0.9997 $\pm$0.0013&  0.0008 $\pm$0.0009 \\ 
MA-proximal & 0.9979 $\pm$0.0037&  3.517 $\pm$0.1829 & 0.9998 $\pm$0.0011&  0.0002 $\pm$0.0004 \\  \hline
\hline
\end{tabular} 
\end{center}
\vspace*{-0.05in}
\caption{\it Comparisons among various approaches: Erd\"{o}s-R\`{e}nyi graph, $n=300, \, 800, \, 3000, \, 6000$, $p=100$, $m=3$. Tuning parameter $\lambda$ picked to yield the highest $F_1$ score. Results based on 100 runs.} \label{table1} \vspace*{-0.15in}
\end{table*}

\begin{figure}[htb]
  \centering
  \includegraphics[width=0.7\linewidth]{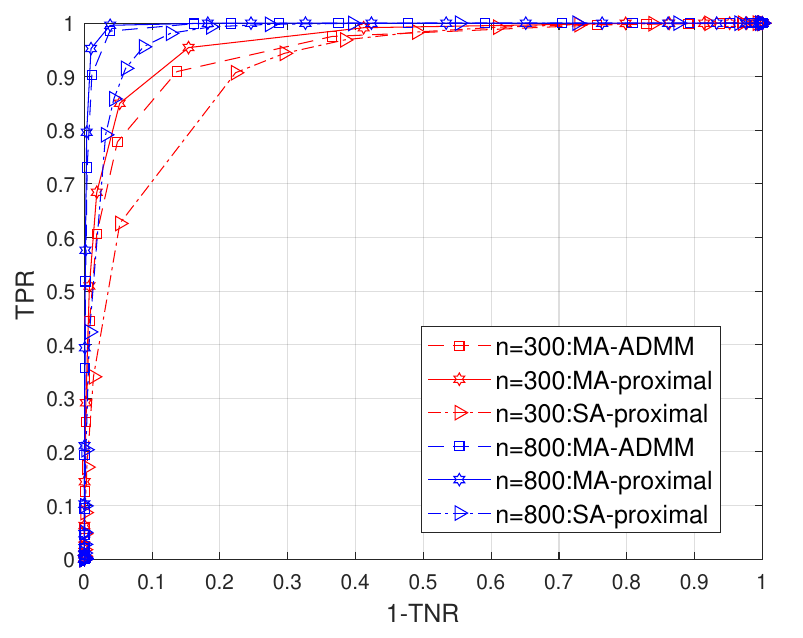} 
	\vspace*{-0.05in}
  \caption{ROC curves for ER graph based on ADMM as well as proximal approaches. }
  \label{fig3}
\end{figure}

\begin{figure}[htb]
  \centering
  \includegraphics[width=0.7\linewidth]{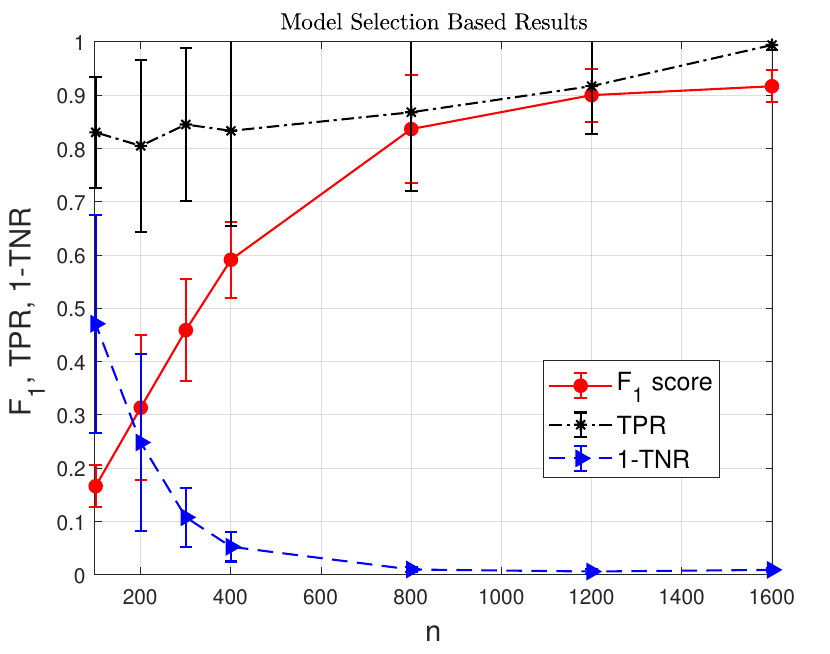} 
	\vspace*{-0.05in}
  \caption{BIC based results for ER graph: $F_1$-scores, TPR and 1-TNR}
  \label{fig4}
\end{figure}

Simulation results based on 100 runs are shown in Figs.\ \ref{fig1}-\ref{fig4}. By changing the penalty parameter $\lambda$ and determining the resulting edges, we calculated the true positive rate (TPR) and false positive rate 1-TNR (where TNR is the true negative rate) over 100 runs. The receiver operating characteristic (ROC) for ER graphs is shown in Fig.\ \ref{fig1} for our MA-ADMM approach (labeled ``MA'') as well as for a SA-ADMM approach (labeled ``SA''), based on \cite{Jiang2018}, where we first estimate an $mp$-node differential graph, and then use $\| \hat{\bm \Delta}^{(k \ell)} \|_F \ne 0$ $\Leftrightarrow$ $\{ \{ k, \ell \} \in {\cal E}_\Delta $. It is seen from Fig.\ \ref{fig1} that our MA-ADMM approach outperforms the SA-ADMM approach (that uses the same cost but element-wise lasso penalty instead of group-lasso penalty). Fig.\ \ref{fig2} is the counterpart of Fig.\ \ref{fig1} for BA graphs., and comments made regarding  Fig.\ \ref{fig1} apply here too.

\vspace*{-0.05in}
\vspace*{-0.05in}
In Fig.\ \ref{fig3} we compare ROC curves of our MA-ADMM approach with that for the MA-proximal and SA-proximal approaches of \cite{Zhao2022} and \cite{Tang2020}, respectively, for $n=300$ and 800. It is seen that the MA-proximal approach outperforms our MA-ADMM approach, while both significantly outperform the SA-proximal approach (and the SA-ADMM approach whose ROC curves are in Fig.\ \ref{fig1}). In Table \ref{table1}, for the ER graph ($n=300, 800$, $p=100$, $m=3$) we compare the four approaches (MA-ADMM, SA-ADMM, MA-proximal, SA-proximal) in terms of the $F_1$ score, execution time (based on tic-toc functions in MATLAB), TPR and 1-TNR, for fixed penalty parameter $\lambda$ selected from a grid of values (the same as for computing the ROC curves) to maximize the $F_1$ score averaged over 100 runs. All algorithms were run on a Window 10 Pro operating system with processor Intel(R) Core(TM) i7-10700 CPU @2.90 GHz with 32 GB RAM, using MATLAB R2023a. Notice that while the MA-proximal approach outperforms our MA-ADMM approach, it also takes more than twice the computation time for the MA-ADMM approach. Similarly, the SA-proximal approach takes more than twice the computation time for the SA-ADMM approach. The latter observation is consistent with the findings of \cite{Tang2020}.

In Table \ref{table1}, we also show results for $n=3000$ and $6000$ for MA-ADMM and MA-proximal approaches in order to provide further empirical validation of the theoretical results stated in Theorem 1. Theorem 1 states that for sufficiently large sample size $n$, one can recover the differential  graph structure exactly w.h.p. It is seen that the $F_1$ score approaches one with increasing $n$, implying graph support recovery, as claimed in Theorem 1(iv). Note that Theorem 1(iv) holds w.h.p., not with probability one, implying a nonzero probability of possibly inexact graph recovery, yielding an $F1$-score less than 1 for $n=3000, 6000$. The sample sizes of $n=300, 800$ are not large enough to yield an $F_1$-score close to 1. There is a lower bound (\ref{eqn352}) on $n$  for Theorem 1(iv) to hold w.h.p. The sample sizes of $n=300, 800$ are apparently less than bound (which is not easily computable since it needs $\alpha$ and $\kappa_\Gamma$).

In Fig.\ \ref{fig4} we show the results based on 100 runs for our approach when BIC parameter selection method (Sec.\ \ref{modelsel}) is applied in conjunction with the MA-ADMM approach. Here we show the TPR, 1-TNR and $F_1$ score values along with the $\pm \sigma$ error bars. The proposed approach works well both in terms of $F_1$ score and TPR vs 1-TNR.   

\begin{figure*}
\begin{subfigure}[b]{0.33\textwidth}
\begin{center}
\includegraphics[width=0.9\linewidth]{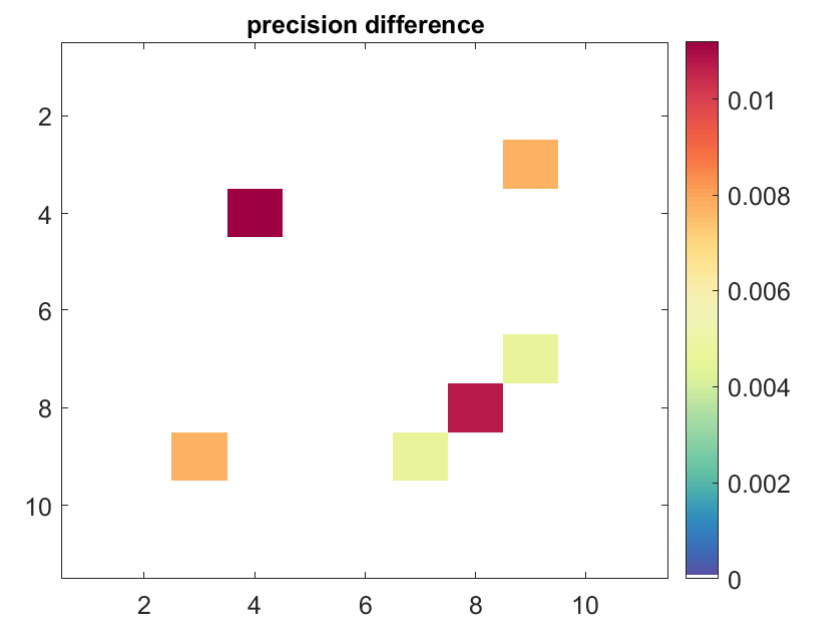}
\caption{Proposed approach: $\| \hat{\bm \Delta}^{(k \ell)} \|_F$, \\ 
\hspace*{0.2in}$k,\ell =1,2, \cdots , 11$}
\end{center}
\end{subfigure}%
\begin{subfigure}[b]{0.33\textwidth}
\begin{center}
\includegraphics[width=0.9\linewidth]{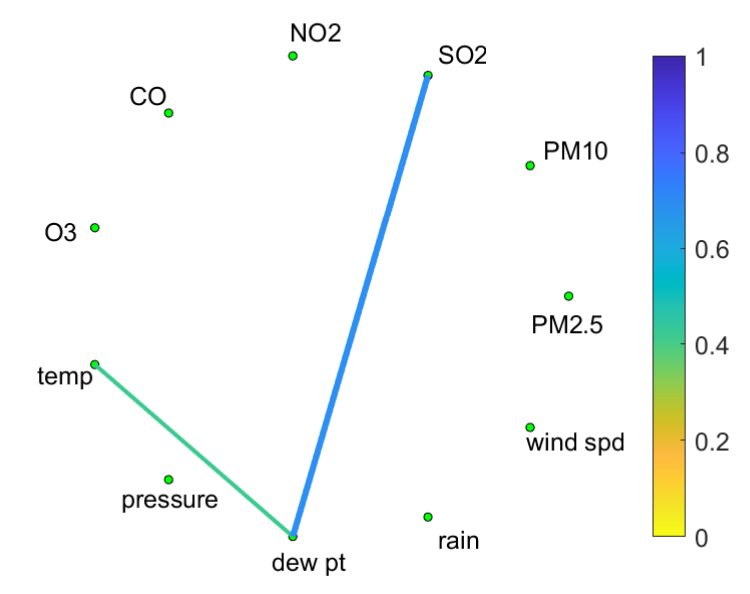}
\caption{Proposed approach: edges \\ 
\hspace*{0.2in}$\big\{ \{k, \ell\} \, : \, \| \hat{\bm \Delta}^{(k \ell)} \|_F > 0 \big\}$}
\end{center}
\end{subfigure}
\begin{subfigure}[b]{0.33\textwidth}
\begin{center}
\includegraphics[width=0.85\linewidth]{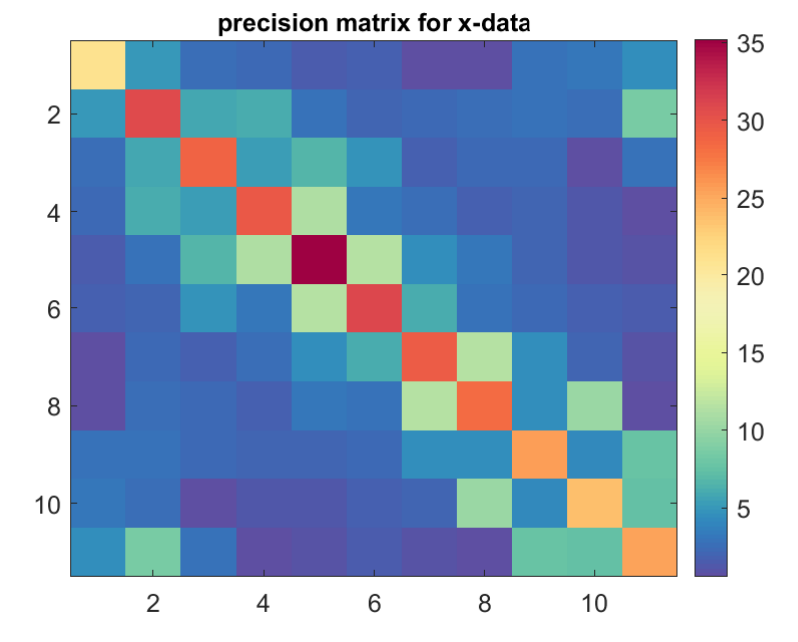}
\caption{$\| \hat{\bm \Omega}_x^{(k \ell)} \|_F$,  $ k,\ell =1,2, \cdots , 11$ \cite{Tugnait21a} \\ $ $}
\end{center}
\vspace*{-0.1in}
\end{subfigure}%
\newline %
\begin{subfigure}[b]{0.33\textwidth}
\begin{center}
\includegraphics[width=0.85\linewidth]{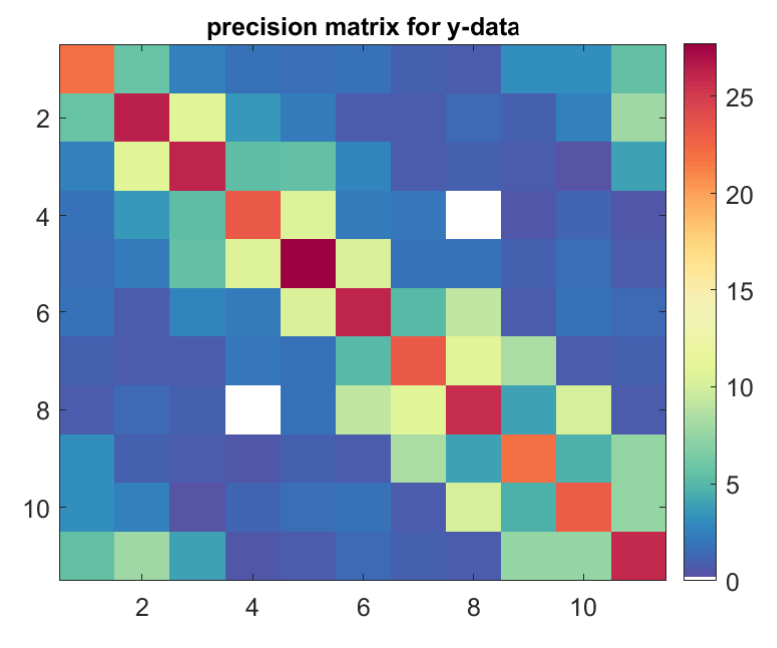}
\caption{$\| \hat{\bm \Omega}_y^{(k \ell)} \|_F$, $k,\ell =1,2, \cdots , 11$ \cite{Tugnait21a} \\ $ $}
\end{center}
\end{subfigure}%
\begin{subfigure}[b]{0.33\textwidth}
\begin{center}
\includegraphics[width=0.85\linewidth]{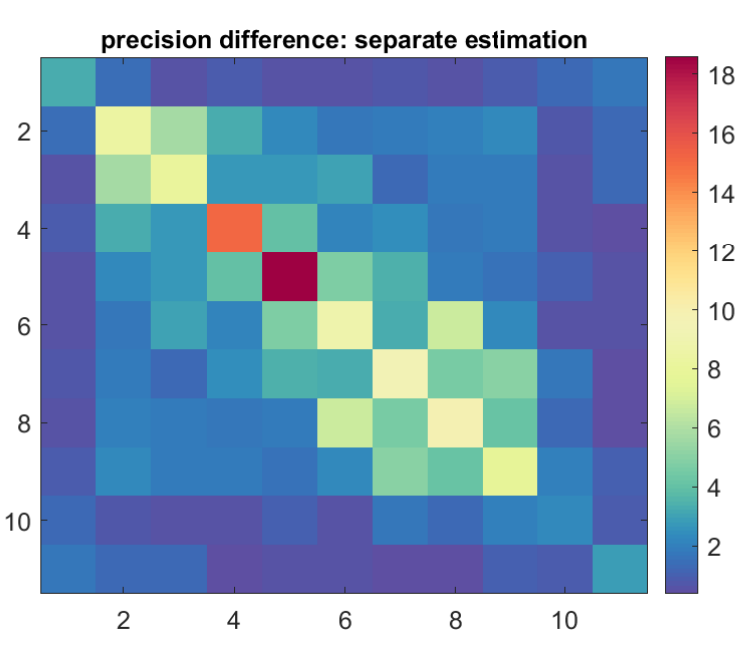}
\caption{$\| \hat{\bm \Omega}_y^{(k \ell)} - \hat{\bm \Omega}_x^{(k \ell)} \|_F$ (\cite{Tugnait21a}) \\ $ $} 
\end{center}
\end{subfigure}%
\begin{subfigure}[b]{0.33\textwidth}
\vspace*{-0.1in}
\begin{center}
\vspace*{-0.1in}
\includegraphics[width=0.9\linewidth]{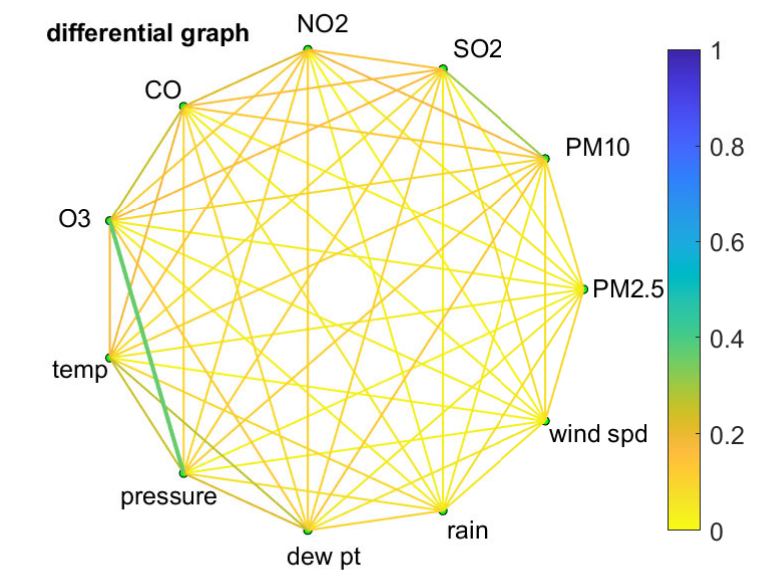}
\caption{Estimated edges \\ \hspace*{0.2in}
$\big\{ \{k,\ell\} \, : \, \| \hat{\bm \Omega}_y^{(k \ell)} - \hat{\bm \Omega}_x^{(k \ell)} \|_F > 0 \big\}$}
\end{center}
\end{subfigure}%
\vspace*{-0.01in}
\caption{Differential graphs comparing Beijing air-quality datasets \cite{Zhang2017} for years 2013-14 and 2014-15: 8 monitoring stations and 11 features ($m=8$, $p=11$, $n_x=n_y=365$). The features are numbered 1-11 beginning PM2.5 (PM$_{2.5}$) and moving counter-clockwise in Fig.\ \ref{fig_real1}(b).} \label{fig_real1}
\end{figure*}

\begin{figure*}
\begin{subfigure}[b]{0.33\textwidth}
\begin{center}
\includegraphics[width=0.8\linewidth]{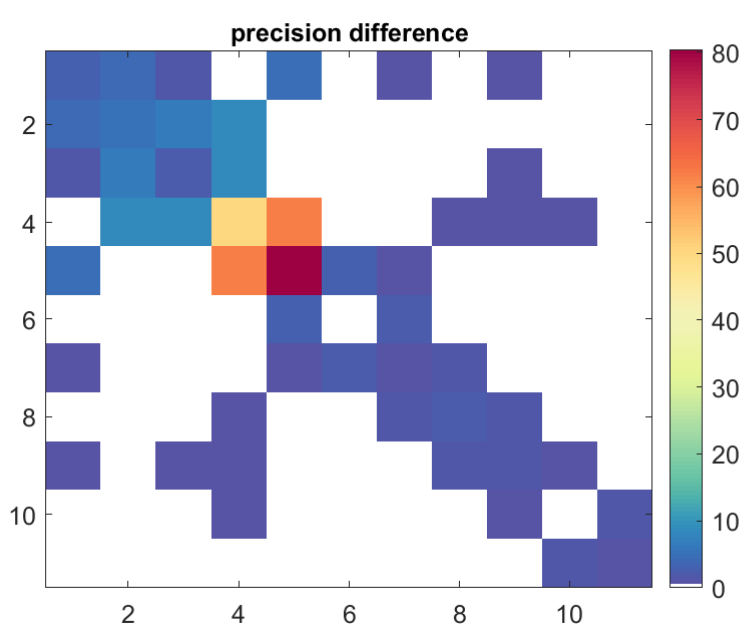}
\caption{Proposed approach: $\| \hat{\bm \Delta}^{(k \ell)} \|_F$, \\ \hspace*{0.2in} $k,\ell =1,2, \cdots , 11$}
\end{center}
\end{subfigure}%
\begin{subfigure}[b]{0.33\textwidth}
\begin{center}
\includegraphics[width=0.9\linewidth]{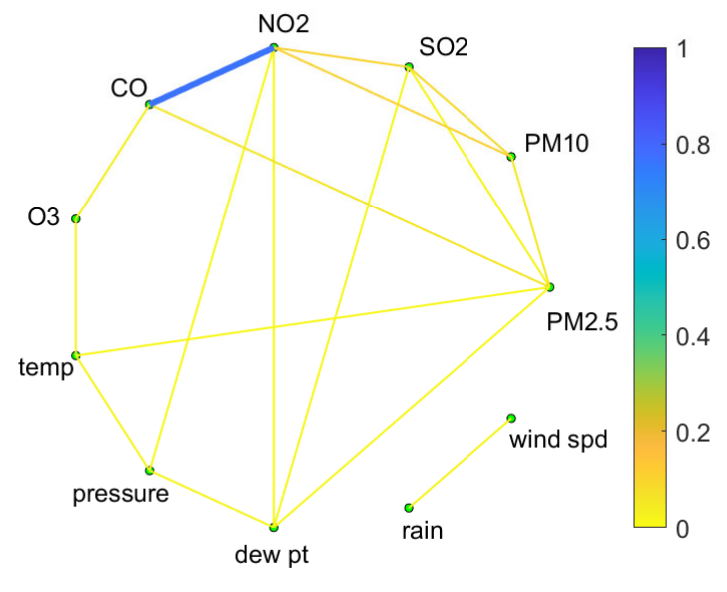}
\caption{Proposed approach: edges \\ \hspace*{0.2in} $\big\{ \{k, \ell\} \, : \, \| \hat{\bm \Delta}^{(k \ell)} \|_F > 0 \big\}$}
\end{center}
\end{subfigure}
\begin{subfigure}[b]{0.33\textwidth}
\begin{center}
\includegraphics[width=0.9\linewidth]{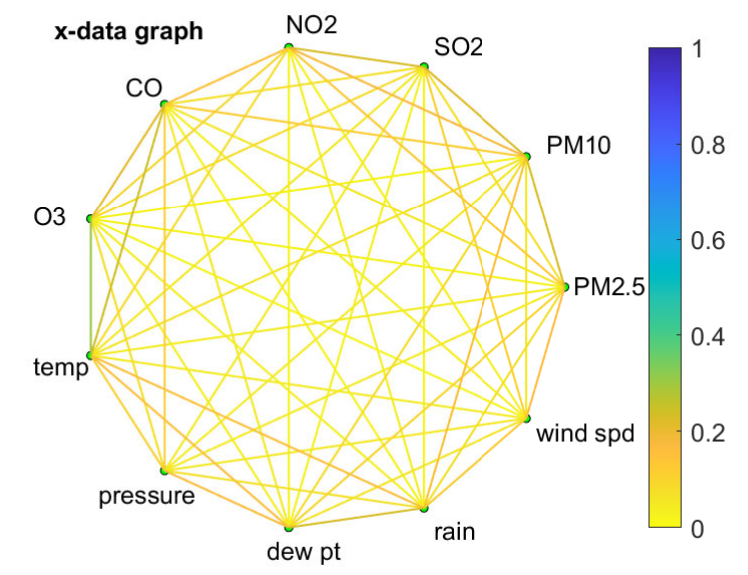}
\caption{Estimated edges \\ \hspace*{0.2in}
$\big\{ \{k,\ell\} \, : \, \| \hat{\bm \Omega}_x^{(k \ell)} \|_F > 0 \big\}$ \cite{Tugnait21a}}
\end{center}
\end{subfigure}%
\newline %
\begin{subfigure}[b]{0.33\textwidth}
\begin{center}
\includegraphics[width=0.9\linewidth]{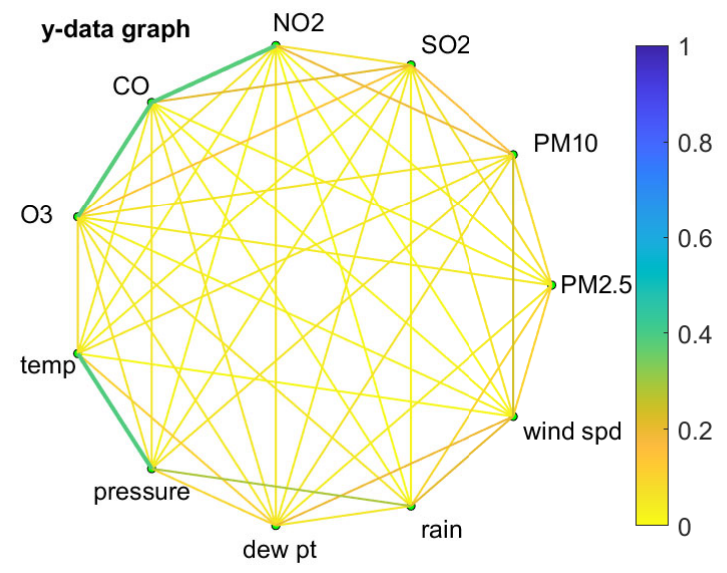}
\caption{Estimated edges \\ \hspace*{0.2in}
$\big\{ \{k,\ell\} \, : \, \| \hat{\bm \Omega}_y^{(k \ell)} \|_F > 0 \big\}$ \cite{Tugnait21a}}
\end{center}
\end{subfigure}%
\begin{subfigure}[b]{0.33\textwidth}
\begin{center}
\includegraphics[width=0.8\linewidth]{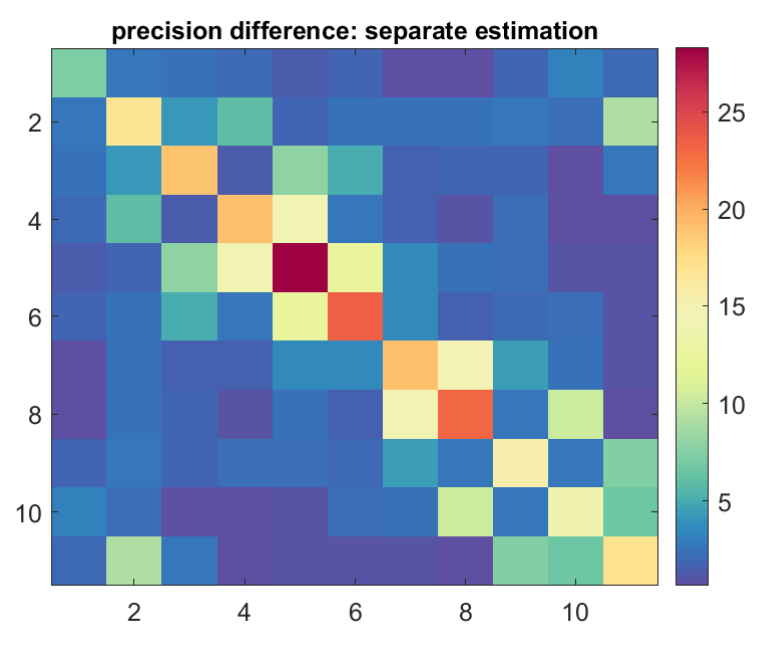}
\caption{$\| \hat{\bm \Omega}_y^{(k \ell)} - \hat{\bm \Omega}_x^{(k \ell)} \|_F$ (\cite{Tugnait21a}) \\ $ $} 
\end{center}
\end{subfigure}%
\begin{subfigure}[b]{0.33\textwidth}
\vspace*{-0.1in}
\begin{center}
\vspace*{-0.1in}
\includegraphics[width=0.9\linewidth]{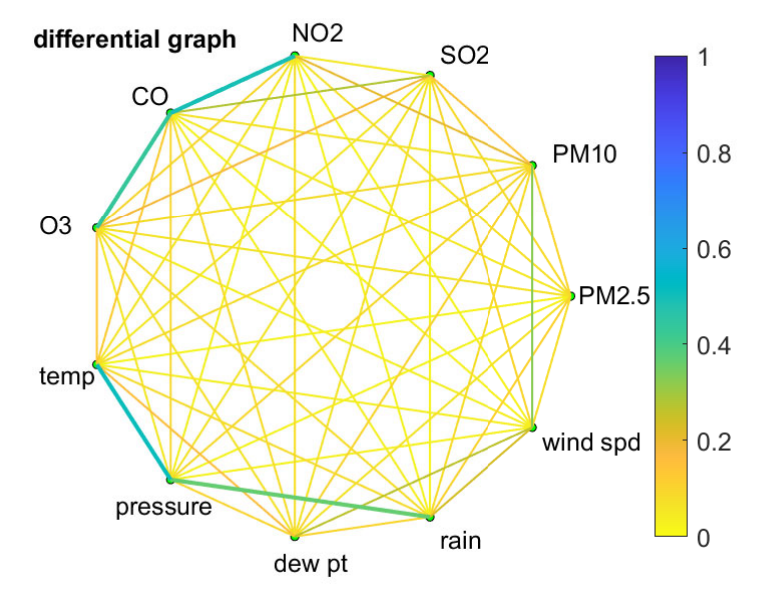}
\caption{Estimated edges \\ \hspace*{0.2in}
$\big\{ \{k,\ell\} \, : \, \| \hat{\bm \Omega}_y^{(k \ell)} - \hat{\bm \Omega}_x^{(k \ell)} \|_F > 0 \big\}$}
\end{center}
\end{subfigure}%
\vspace*{-0.01in}
\caption{Differential graphs comparing Beijing air-quality datasets \cite{Zhang2017} acquired from two sets of monitoring stations, 4 stations per set, year 2013-14: 4 monitoring stations and 11 features ($m=4$, $p=11$, $n_x=n_y=365$).} \label{fig_real2}
\end{figure*}
\vspace*{-0.2in}

\subsection{Real Data: Beijing air-quality dataset \cite{Zhang2017}} \label{NEreal}
Here we consider Beijing air-quality dataset \cite{Zhang2017, Chen2015}, downloaded from \url{https://archive.ics.uci.edu/ml/datasets/Beijing+Multi-Site+Air-Quality+Data}. This data set includes hourly air pollutants data from 12 nationally-controlled air-quality monitoring sites in the Beijing area from the Beijing Municipal Environmental Monitoring Center, and meteorological data in each air-quality site are matched with the nearest weather station from the China Meteorological Administration. The time period is from March 1st, 2013 to February 28th, 2017.The six air pollutants are PM$_{2.5}$, PM$_{10}$, SO$_2$, NO$_2$, CO, and O$_3$, and the meteorological data is comprised of five features: temperature, atmospheric pressure, dew point, wind speed, and rain; we did not use wind direction. Thus we have eleven features. We used data from 8 sites: 4 suburban/rural sites -- Changping, Huairou, Shunyi, Dingling,  and 4 urban area stations -- Aotizhongxin, Dongsi, Guanyuan, Gucheng \cite[Fig.\ 1]{Chen2015}. The data are averaged over 24 hour period to yield daily averages. We used one year of daily data resulting in $n_x = n_y = 365$ days. The stations are used as attributes, with $m=8$ for comparison between years 2013-14 and 2014-15, and $m=4$ for comparison between suburban/rural sites and urban sites using 2013-14 year data.

We pre-process the data as follows. Given $i$th feature data ${\bm z}_i(t) \in \mathbb{R}^m$, we transform it to $\bar{\bm z}_i(t) = \ln({\bm z}_i(t)/{\bm z}_i(t-1))$ and then detrend it (i.e., remove the best straight-line fit using the MATLAB function detrend). Finally, we scale the detrended scalar sequence to have a mean-square value of one over $n_x$ or $n_y$ samples. The logarithmic transformation and detrending of each feature sequence makes the sequence closer to (univariate) stationary and Gaussian, while scaling ``balances'' the possible wide variations in the scale of various feature measurements. All temperatures were converted from Celsius to Kelvin to avoid negative numbers, and if a value of a feature is zero (e.g., wind speed), we added a small positive number to it, so that the logarithmic transformation is well-defined. 

Fig.\ \ref{fig_real1} shows the estimated differential graphs when comparing daily-averaged data from 2013-14 ($x$-data) to that from 2014-15 ($y$-data), with air-quality and meteorological variables as $p=11$ features measured at 8 monitoring sites ($m$=8). The objective is to visualize and explore differential conditional dependency relationships among the 11 variables, comparing one year to another, to investigate if pollution reduction measures have had any impact. Our intuition is that one does not expect such rapid changes within a short period of one year (see also \cite{Zhang2017}), therefore, our method should confirm our intuition. Figs.\ \ref{fig_real1}(a)-(b) show estimated $\| \hat{\bm \Delta}^{(k \ell)} \|_F$ for various edges $\{k, \ell \}$, where it is unscaled in Fig.\ \ref{fig_real1}(a) but scaled in Fig.\ \ref{fig_real1}(b) so that the largest $\| \hat{\bm \Delta}^{(k \ell)} \|_F$ (including $k=\ell$) is normalized to one. It is seen that differential graph weights are essentially zero (very sparse), implying that there are no year-to-year changes in the conditional dependency relationships among the 11 variables. This observation conforms to the findings of \cite{Zhang2017, Chen2015}: no significant year-to-year changes. We also estimated the MA graphs for each year separately as $\| \hat{\bm \Omega}_x^{(k \ell)} \|_F$ and $\| \hat{\bm \Omega}_y^{(k \ell)} \|_F$ (shown in Figs.\ \ref{fig_real1}(c)-(d)), using the approach of \cite{Tugnait21a}, and based on the individual estimates, we computed the differential graph $\| \hat{\bm \Omega}_y^{(k \ell)} - \hat{\bm \Omega}_x^{(k \ell)} \|_F$ (shown in Figs.\ \ref{fig_real1}(e)-(f)). It is seen the separate-estimation approach does not yield a sparse differential graph, even though the two individual graphs in Figs.\ \ref{fig_real1}(c)-(d) are not all that different.

Fig.\ \ref{fig_real2} shows the estimated differential graphs when comparing daily-averaged data over the period 2013-14, from four suburban/rural sites ($x$-data) to that from four urban sites ($y$-data), with air-quality and meteorological variables as $p=11$ features measured at two sets of 4 monitoring sites ($m$=4). The objective again is to visualize and explore differential conditional dependency relationships among the 11 variables, but in this case comparing one subregion to another. There are significant differences in meteorological conditions and pollutant sources, levels and mutual interactions, among suburban and urban areas \cite{Zhang2017, Chen2015}. The suburban areas (located toward north) are less polluted than the urban areas (located toward south) \cite{Zhang2017, Chen2015}. Automobile exhaust is the main cause of NO$_2$ which is likely to undergo a chemical reaction with Ozone O$_3$, thereby, lowering its concentration \cite{Chen2015}. Cold, dry air from the north reduces both dew point and PM$_{2.5}$ particle concentration in suburban areas while southerly wind brings warmer and more humid air from the more polluted south that elevates the PM$_{2.5}$ concentration \cite{Zhang2017}. The urban stations neighbor the south of Beijing which is heavily installed with iron, steel and cement industries in Hebei province \cite{Zhang2017}. Figs.\ \ref{fig_real2}(a)-(b) show estimated $\| \hat{\bm \Delta}^{(k \ell)} \|_F$ for various edges $\{k, \ell \}$, where it is unscaled in Fig.\ \ref{fig_real2}(a) but scaled in Fig.\ \ref{fig_real2}(b) so that the largest $\| \hat{\bm \Delta}^{(k \ell)} \|_F$ (including $k=\ell$) is normalized to one. It is seen that quite a few of the differential graph weights are significantly non-zero in Fig.\ \ref{fig_real2}(a), unlike that in Fig.\ \ref{fig_real1}(a), implying significant differences in the conditional dependency relationships among the 11 variables for suburban and urban areas. This observation conforms to the findings of \cite{Zhang2017, Chen2015}. The comments made regarding Figs.\ \ref{fig_real1}(c)-(f) apply as well to Figs.\ \ref{fig_real2}(c)-(f).

\vspace*{-0.2in}
\section{Conclusions} 
\vspace*{-0.05in}
A group lasso penalized D-trace loss function approach for differential graph learning from multi-attribute data was presented. An ADMM algorithm was presented to optimize the convex objective function. Theoretical analysis establishing consistency of the estimator in high-dimensional settings was performed. We tested the proposed approach on synthetic as well as real data. In the synthetic data example, the multi-attribute approach is shown to outperform a single-attribute approach in correctly detecting the differential graph edges with ROC as the performance metric.
\vspace*{-0.2in}

\appendices
\section{Technical Lemmas and Proof of Theorem 1} \label{append1}
In this Appendix, we provide a proof of Theorem 1. 
 A necessary and sufficient condition for minimization of convex $L_\lambda({\bm \Delta}, \hat{\bm \Sigma}_x , \hat{\bm \Sigma}_y)$ given by (\ref{eqn20}) w.r.t.\ ${\bm \Delta} \in \mathbb{R}^{mp \times mp}$ is that $\hat{\bm \Delta}$ minimizes (\ref{eqn20}) iff the zero matrix belongs to the sub-differential of $L_\lambda({\bm \Delta}, \hat{\bm \Sigma}_x , \hat{\bm \Sigma}_y)$. That is,
\begin{align} 
  {\bm 0} = & \frac{\partial L({\bm \Delta}, \hat{\bm \Sigma}_x , \hat{\bm \Sigma}_y)}{\partial {\bm \Delta}}
	     + \lambda {\bm Z}({\bm \Delta}) \, \Big|_{{\bm \Delta} = \hat{\bm \Delta}} \nonumber \\
			= & \hat{\bm \Sigma}_x \hat{\bm \Delta} \hat{\bm \Sigma}_y - (\hat{\bm \Sigma}_x-\hat{\bm \Sigma}_y) 
			           + \lambda {\bm Z}(\hat{\bm \Delta})
  \label{aeqn210}  
\end{align}
where ${\bm Z}({\bm \Delta}) \in \partial \sum_{k, \ell=1}^p \| {\bm \Delta}^{(k \ell)} \|_F \in \mathbb{R}^{mp \times mp}$, the sub-differential of group lasso penalty term, is given by \cite{Friedman2010a, Friedman2010b}
\begin{align} 
  & ({\bm Z}({\bm \Delta}))^{(k \ell)}   \nonumber \\
	=& \left\{ \begin{array}{ll} 
	      \frac{{\bm \Delta}^{(k \ell)}}{\| {\bm \Delta}^{(k \ell)} \|_F} & \mbox{if }\, 
				                  \| {\bm \Delta}^{(k \ell)} \|_F \ne 0 \\
				{\bm V} \in \mathbb{R}^{m \times m} , \; \| {\bm V} \|_F \le 1, & \mbox{if } \, 
				                  \| {\bm \Delta}^{(k \ell)} \|_F = 0 \end{array} \right. \, .
  \label{aeqn215}  
\end{align}
In terms of $m \times m$ submatrices of ${\bm \Delta}$, $\hat{\bm \Sigma}_x$, $\hat{\bm \Sigma}_y$ and ${\bm Z}({\bm \Delta})$ corresponding to various graph edges, using $\mbox{bvec}({\bm A} {\bm D} {\bm B}) = ({\bm B}^\top \boxtimes {\bm A}) \mbox{bvec}({\bm D})$ \cite[Lemma 1]{Tracy1989}, we may rewrite (\ref{aeqn210}) as
\begin{align} 
   (\hat{\bm \Sigma}_y \boxtimes  \hat{\bm \Sigma}_x) \mbox{bvec}(\hat{\bm \Delta} )  - \mbox{bvec}(\hat{\bm \Sigma}_x-\hat{\bm \Sigma}_y) + \lambda \, \mbox{bvec}({\bm Z}(\hat{\bm \Delta}))   = {\bm 0}
  \label{aeqn220}  
\end{align}
Then (\ref{aeqn220}) can be rewritten as
\begin{align} 
  &  \begin{bmatrix} \hat{\bm \Gamma}_{S,S} & \hat{\bm \Gamma}_{S,S^c} \\
	   \hat{\bm \Gamma}_{S^c,S} & \hat{\bm \Gamma}_{S^c,S^c} \end{bmatrix} 
		 \begin{bmatrix} \mbox{bvec}(\hat{\bm \Delta}_S ) \\ \mbox{bvec}(\hat{\bm \Delta}_{S^c} ) \end{bmatrix}
	-	\begin{bmatrix} \mbox{bvec}((\hat{\bm \Sigma}_x-\hat{\bm \Sigma}_y)_S) \\ 
	  \mbox{bvec}((\hat{\bm \Sigma}_x-\hat{\bm \Sigma}_y)_{S^c} ) \end{bmatrix}  \nonumber \\
	& \quad	+ \lambda \, \begin{bmatrix} \mbox{bvec}({\bm Z}(\hat{\bm \Delta}_S)) \\
			   \mbox{bvec}({\bm Z}(\hat{\bm \Delta}_{S^c})) \end{bmatrix} = 
				\begin{bmatrix} {\bm 0} \\ {\bm 0} \end{bmatrix} \, . \label{aeqn227}  
\end{align}

The general approach of \cite{Ravikumar2011} (followed in \cite{Kolar2014, Zhang2014, Yuan2017, Jiang2018}) is to first solve the hypothetical constrained optimization problem with known edgeset $S$
\begin{equation}
  \tilde{\bm \Delta} = \arg\min_{{\bm \Delta}: {\bm \Delta}_{S^c} 
	 = {\bm 0}} L_\lambda({\bm \Delta}, \hat{\bm \Sigma}_x , \hat{\bm \Sigma}_y)  \label{aeqn250}
\end{equation}
where $S^c$ is the complement of $S$. 
Since, by construction, $\tilde{\bm \Delta}_{S^c} = {\bm 0}$, in this case (\ref{aeqn227}) reduces to 
\begin{align} 
   & \hat{\bm \Gamma}_{S,S} \mbox{bvec}(\tilde{\bm \Delta}_S )  - \mbox{bvec}((\hat{\bm \Sigma}_x-\hat{\bm \Sigma}_y)_S) + \lambda \, \mbox{bvec}({\bm Z}(\tilde{\bm \Delta}_S))    = {\bm 0} \, . \label{aeqn260}  
\end{align}
In the approach of \cite{Ravikumar2011}, one investigates conditions under which the solution $\hat{\bm \Delta}$ to (\ref{eqn20}) is the same as the solution $\tilde{\bm \Delta}$ to (\ref{aeqn250}). This is done by showing that $\hat{\bm \Delta}$ satisfies (\ref{aeqn227}). The choice $\hat{\bm \Delta} =\tilde{\bm \Delta}$ implies that $\hat{\bm \Delta}_{S^c} = {\bm 0}$ and (\ref{aeqn260}) is true with $\tilde{\bm \Delta}$ replaced with $\hat{\bm \Delta}$. In order to satisfy (\ref{aeqn227}), it remains to show that for any edge $e \in S^c$, we have strict feasibility
\begin{align} 
 \| \hat{\bm \Gamma}_{e,S} \mbox{bvec}(\tilde{\bm \Delta}_S )  - \mbox{bvec}((\hat{\bm \Sigma}_x-\hat{\bm \Sigma}_y)_e) \|_2
	       < \lambda \, , \label{aeqn261}
\end{align}
where for ${\bm a} \in \mathbb{R}^q$, $\|{\bm a} \|_2 = \sqrt{{\bm a}^\top {\bm a}}$. This requires a set of sufficient conditions stated in Theorem 1. 

Lemma 1 follows from \cite[Lemma 1]{Ravikumar2011}. It is stated for more general sub-Gaussian distributions as in \cite[Lemma 1]{Ravikumar2011}, but will be used later for Gaussian distributions, a subset of sub-Gaussian distributions. \\
{\bf Lemma 1}. Suppose $\hat{\bm \Sigma} = (1/n) \sum_{t=1}^n {\bm x}(t) {\bm x}^\top (t)$, given $n$ i.i.d.\ samples $\{ {\bm x}(t) \}_{t=1}^n$ of zero-mean sub-Gaussian ${\bm x} \in \mathbb{R}^{mp}$ with covariance ${\bm \Sigma}^\ast$ such that each component $x_i/\sqrt{\Sigma_{ii}^\ast}$ is sub-Gaussian with parameter $\sigma_{sg}$. Define 
$\sigma_{max} = \max_{1 \le i \le mp_n} \Sigma_{ii}^\ast$ and 
\begin{equation}  \label{aeqn300}
 \tilde{C}_0 = 8 (1 + 4 \sigma_{sg}^2) m \sigma_{max}  \sqrt{2 \big(\tau + \ln(4m^2)/ \ln(p_n) \big)}.
\end{equation}
Then
\begin{equation}  \label{aeqn305}
   P \Big(  \| \bm{\mathcal C}(\hat{\bm \Sigma}- {\bm \Sigma}^\ast) \|_\infty
	    > \tilde{C}_0 \sqrt{\ln(p_n) / n} \Big) \le 1/p_n^{\tau -2} 
\end{equation}
for any $\tau > 2$ and $n > 2m^2 (\ln(4m^2) + \tau \ln( p_n))$. $\quad \bullet$ \\
{\it Proof}. By \cite[Lemma 1]{Ravikumar2011}, with $b = 8 (1 + 4 \sigma_{sg}^2)$, we have
\begin{equation}  \label{aeqn3051}
   P \Big(  | [\hat{\bm \Sigma}- {\bm \Sigma}^\ast ]_{ij} | > \delta \Big)
	    \le 4 \, \exp(-c_\ast n \delta^2 )
\end{equation}
for any $\delta \in (0, b \, \sigma_{max} )$ where $c_\ast^{-1} = 2 b^2 \sigma_{max}^2$. For any edge $\{k, \ell\}$ of the MA graph, with $m^2$ edges $\{i,j\}$ of the corresponding SA graph associated with $\{k, \ell\}$, using the union bound, we have
\begin{align} 
   P & \Big(  | [\bm{\mathcal C}(\hat{\bm \Sigma}- {\bm \Sigma}^\ast)]_{kl} | > \delta \Big) \nonumber \\
	  &  \le P \Big( \max_{\{i,j\} \in \{k, \ell \}} ( [\hat{\bm \Sigma}- {\bm \Sigma}^\ast ]_{ij})^2 
		 > \frac{\delta^2}{m^2} \Big) \nonumber \\
		& \le m^2 P \Big(  | [\hat{\bm \Sigma}- {\bm \Sigma}^\ast ]_{ij} | 
		        > \frac{\delta}{m} \Big)  
	 = 4 m^2 \exp \big(-c_\ast n \frac{\delta^2}{m^2} \big) \, . \label{aeqn3052}  
\end{align}
Applying the union bound once more over all $p_n^2$ entries
\begin{align} 
   P & \Big(  \| \bm{\mathcal C}(\hat{\bm \Sigma}- {\bm \Sigma}^\ast) \|_\infty > \delta \Big) 
	   \le 4 (mp_n)^2 \exp \big(-c_\ast n \frac{\delta^2}{m^2} \big) =: P_{tb} \, . \label{aeqn3054}  
\end{align}
Choose $\delta = \tilde{C}_0 \sqrt{\ln(p_n) / n} = b m \sigma_{max} \sqrt{2 \ln(4p_n^\tau m^2) / n} \,$. Then we have 
\begin{align} 
   P_{tb} = & 4 (mp_n)^2 \exp \big(\ln(4 p_n^\tau m^2)^{-1} \big) = 1/p_n^{\tau -2}  \label{aeqn3055}  
\end{align}
provided $\delta \in (0, b \sigma_{max} )$. Therefore, we need to have $\tilde{C}_0 \sqrt{\ln(p_n) / n} < b \sigma_{max}$ requiring $n > 2m^2 (\ln(4m^2) + \tau \ln( p_n))$. This completes the proof. $\quad \Box$  

Using the union bound, Lemma 1 and Gaussian assumption, we have Lemma 2.
\\
{\bf Lemma 2}. Let $\hat{\bm \Sigma}_x$ and $\hat{\bm \Sigma}_y$ be as in (\ref{eqn10}), $\bar{\sigma}_{xy}$ as in (\ref{eqn336}), ${C}_0$ as in (\ref{eqn337}) and assume data are Gaussian. Define $n = \min(n_x, n_y)$ and
\begin{align} 
   {\mathcal A} =  & \max \big\{ \| \bm{\mathcal C}(\hat{{\bm \Sigma}_x}-{\bm \Sigma}_x^\ast) \|_\infty \, , 
	 \| \bm{\mathcal C}(\hat{{\bm \Sigma}_y}-{\bm \Sigma}_y^\ast) \|_\infty \big\}  \, . \label{aeqn310}
\end{align}
Then for any $\tau > 2$ and $n > 2m^2 \ln(4m^2 p_n^\tau)$,
\begin{align}  
   P & \Big(  {\mathcal A}  > {C}_0 \sqrt{\ln(p_n)/n} \Big) \le 2/p_n^{\tau -2} \quad \bullet \label{aeqn315}
\end{align}
{\it Proof}. For Gaussian distribution, the sub-Gaussian parameter $\sigma_{sg}$ of Lemma 1 equals 1. Then $8 (1 + 4 \sigma_{sg}^2) = 40$. Let $C_{0x} = 40 m (\max_i \Sigma_{x,ii}^\ast)  \sqrt{2 \big(\tau + \ln(4m^2)/ \ln(p_n) \big)}$ and $C_{0y} = 40 m (\max_i \Sigma_{y,ii}^\ast)  \sqrt{2 \big(\tau + \ln(4m^2)/ \ln(p_n) \big)}$ (where $\Sigma_{x,ii}^\ast =[{\bm \Sigma}_x^\ast)]_{ii}$, etc.). Then using Lemma 1 and union bound,
\begin{align}  
   P & \Big(  {\mathcal A}  > {C}_0 \sqrt{\ln(p_n)/n} \Big) \nonumber \\
	  \le & P \Big(  \| \bm{\mathcal C}(\hat{\bm \Sigma}_x- {\bm \Sigma}_x^\ast) \|_\infty
	    > C_0 \sqrt{\ln(p_n) / n} \Big) \nonumber \\ 
			& \; + P \Big(  \| \bm{\mathcal C}(\hat{\bm \Sigma}_y- {\bm \Sigma}_y^\ast) \|_\infty
	    > C_0 \sqrt{\ln(p_n) / n} \Big)  \nonumber \\
	  \le & 2/p_n^{\tau -2}  \label{aeqn316}
\end{align}
since $C_0 \ge C_{0x}$ and $C_0 \ge C_{0y}$. $\quad \Box$

Recall (\ref{eqn225})-(\ref{eqn335}) and define 
\begin{align}
  {\bm \Delta}_x = & \hat{\bm \Sigma}_x -{\bm \Sigma}_x^\ast \, , \;
	{\bm \Delta}_y = \hat{\bm \Sigma}_y -{\bm \Sigma}_y^\ast \, , \;  
	{\bm \Delta}_\Gamma =  \hat{\bm \Gamma} -{\bm \Gamma}^\ast \, , \label{aeqn400} \\
	{\bm \Delta}_\Sigma = & {\bm \Delta}_x - {\bm \Delta}_y \, , \;
	{\epsilon}_x =  \|\bm{\mathcal C}({\bm \Delta}_x)\|_\infty \, , \label{aeqn401} \\ 
		{\epsilon}_y = & \|\bm{\mathcal C}({\bm \Delta}_y)\|_\infty \, , \;
		{\epsilon} > \max\{ {\epsilon}_x, {\epsilon}_y \} \,.  \label{aeqn403}	
\end{align}
{\bf Lemma 3}.  Assume that
\begin{align}
  \kappa_\Gamma &  < \frac{1}{3 s_n (\epsilon^2+2M \epsilon)}  \, .
    \label{aeqn405}	
\end{align}
Let (${\bm \Gamma}_{S,S}^{-\ast}$ denotes $({\bm \Gamma}_{S,S}^\ast)^{-1}$)
\begin{align}
   R({\bm \Delta}_\Gamma)= & \hat{\bm \Gamma}_{S,S}^{-1} - {\bm \Gamma}_{S,S}^{-\ast}  
	  + {\bm \Gamma}_{S,S}^{-\ast} ({\bm \Delta}_\Gamma)_{S,S} {\bm \Gamma}_{S,S}^{-\ast}\, .
    \label{aeqn406}	
\end{align}
Then we have
\begin{align}
  & \| \bm{\mathcal C}( R({\bm \Delta}_\Gamma)) \|_\infty \le  
	\frac{3}{2} \kappa_\Gamma^3 s_n ( \epsilon^2 + 2M \epsilon)^2 \, ,
    \label{aeqn407}	\\
	& \| \bm{\mathcal C}( R({\bm \Delta}_\Gamma)) \|_{1,\infty} \le  
	   \frac{3}{2} \kappa_\Gamma^3 s_n^2 ( \epsilon^2 + 2M \epsilon)^2 \, ,
    \label{aeqn408}	\\
 & \| \bm{\mathcal C}( \hat{\bm \Gamma}_{S,S}^{-1} - {\bm \Gamma}_{S,S}^{-\ast} ) \|_\infty \nonumber \\
  & \;\; \le  
	 \kappa_\Gamma^2 (\epsilon^2+2M \epsilon) \big(1+ 1.5 s_n \kappa_\Gamma (\epsilon^2+2M \epsilon) \big) \, ,
    \label{aeqn409}	\\
	&	\| \bm{\mathcal C}( \hat{\bm \Gamma}_{S,S}^{-1} - {\bm \Gamma}_{S,S}^{-\ast} ) \|_{1,\infty} \le 
	  s_n \| \bm{\mathcal C}( \hat{\bm \Gamma}_{S,S}^{-1} - {\bm \Gamma}_{S,S}^{-\ast} ) \|_\infty \quad \bullet
    \label{aeqn410}
\end{align}
{\it Proof}. We have 
\begin{align}
	{\bm \Delta}_\Gamma = & \hat{\bm \Sigma}_y \boxtimes \hat{\bm \Sigma}_x - 
	           {\bm \Sigma}_y^\ast \boxtimes {\bm \Sigma}_x^\ast \nonumber \\
	= & {\bm \Delta}_y \boxtimes {\bm \Delta}_x + {\bm \Sigma}_y^\ast \boxtimes {\bm \Delta}_x 
	  + {\bm \Delta}_y \boxtimes {\bm \Sigma}_x^\ast . \label{aeqn420}
\end{align}
By \cite[Lemma 14]{Kolar2014}, 
\begin{align}
	\| \bm{\mathcal C}( {\bm A} {\bm B} ) \|_{1,\infty} & \le \| \bm{\mathcal C}( {\bm A}  ) \|_{1,\infty}
	  \| \bm{\mathcal C}( {\bm B} ) \|_{1,\infty} \label{aeqn423}
\end{align}
and by \cite[Lemma 15]{Kolar2014}, 
\begin{align}
	\| \bm{\mathcal C}( {\bm A} {\bm B} ) \|_{\infty} & \le \| \bm{\mathcal C}( {\bm A}  ) \|_{\infty}
	  \| \bm{\mathcal C}( {\bm B}^\top ) \|_{1,\infty} \, . \label{aeqn425}
\end{align}
Since $\|{\bm A} \otimes {\bm B} \|_F = \|{\bm A} \|_F \, \| {\bm B} \|_F$ and ${\bm A} \boxtimes {\bm B} = [{\bm A}_{ij} \boxtimes {\bm B}]_{ij} = [[{\bm A}_{ij} \otimes {\bm B}_{k \ell}]_{k \ell} ]_{ij}$, we have 
\begin{align}
	\| \bm{\mathcal C}( {\bm A} \boxtimes {\bm B} ) \|_{\infty} & \le \| \bm{\mathcal C}( {\bm A}  ) \|_{\infty}
	  \| \bm{\mathcal C}( {\bm B} ) \|_{\infty} \, . \label{aeqn427}
\end{align}
From (\ref{eqn320}), (\ref{aeqn403}), (\ref{aeqn420}) and (\ref{aeqn427}),
\begin{align}
	\| \bm{\mathcal C}({\bm \Delta}_\Gamma ) \|_{\infty} & \le 
	  {\epsilon}_x {\epsilon}_y + M {\epsilon}_x + M {\epsilon}_y
		< \epsilon^2 + 2M \epsilon \label{aeqn428}
\end{align} 
and since $|S| = s_n$,
\begin{align}
	\| \bm{\mathcal C}(({\bm \Delta}_\Gamma)_{S,S} ) \|_{1,\infty} & \le 
	  s_n \| \bm{\mathcal C}(({\bm \Delta}_\Gamma)_{S,S} ) \|_{\infty}
		  \le s_n \| \bm{\mathcal C}({\bm \Delta}_\Gamma ) \|_{\infty} \nonumber \\
		& < s_n( \epsilon^2 + 2M \epsilon) \, . \label{aeqn4290}
\end{align}
By assumption (\ref{aeqn405}),
\begin{align}
	\kappa_\Gamma \| \bm{\mathcal C}(({\bm \Delta}_\Gamma)_{S,S} ) \|_{1,\infty} & = 
	  \| \bm{\mathcal C}( (\Gamma_{S,S}^\ast)^{-1} ) \|_{1,\infty}  
		    \| \bm{\mathcal C}(({\bm \Delta}_\Gamma)_{S,S} ) \|_{1,\infty} \nonumber \\
		& < \frac{1}{3} \, . \label{aeqn429}
\end{align}
By (\ref{aeqn429}) we can invoke \cite[Lemma 5]{Ravikumar2011} to have
\begin{align}
	R({\bm \Delta}_\Gamma) = &  {\bm \Gamma}_{S,S}^{-\ast} ({\bm \Delta}_\Gamma)_{S,S} {\bm \Gamma}_{S,S}^{-\ast}
	   ({\bm \Delta}_\Gamma)_{S,S} {\bm J} {\bm \Gamma}_{S,S}^{-\ast} \label{aeqn430}
\end{align}
where ${\bm J} = \sum_{k=0}^\infty (-1)^k \big( {\bm \Gamma}_{S,S}^{-\ast} ({\bm \Delta}_\Gamma)_{S,S} \big)^k$.
Using (\ref{aeqn423}), (\ref{aeqn425}) and (\ref{aeqn430}), we have
\begin{align}
&	\| \bm{\mathcal C}( R({\bm \Delta}_\Gamma)) \|_\infty \; \le  \; 
	 \| \bm{\mathcal C}( {\bm \Gamma}_{S,S}^{-\ast} ({\bm \Delta}_\Gamma)_{S,S} ) \|_\infty  \nonumber \\
	  & \quad\quad\quad\quad \times\| \bm{\mathcal C}({\bm \Gamma}_{S,S}^{-\ast}
	   ({\bm \Delta}_\Gamma)_{S,S} J {\bm \Gamma}_{S,S}^{-\ast} )^\top \|_{1,\infty} \nonumber \\
	  & \le \| \bm{\mathcal C}( {\bm \Gamma}_{S,S}^{-\ast}) \|_{1,\infty}^3  
		 \| \bm{\mathcal C}(  ({\bm \Delta}_\Gamma)_{S,S} ) \|_\infty 
		 \| \bm{\mathcal C}(  ({\bm \Delta}_\Gamma)_{S,S} ) \|_{1,\infty} \nonumber \\
	  & \quad\quad\quad\quad  \times \| \bm{\mathcal C}( {\bm J}^\top ) \|_{1,\infty}  \, . \label{aeqn435}
\end{align}
Now using (\ref{aeqn429}),
\begin{align}
&	\| \bm{\mathcal C}( {\bm J}^\top ) \|_{1,\infty} \; \le  \; 
	 \sum_{k=0}^\infty \| \bm{\mathcal C}( {\bm \Gamma}_{S,S}^{-\ast} ) \|_{1,\infty}^k 
	                \| \bm{\mathcal C}( ({\bm \Delta}_\Gamma)_{S,S} ) \|_{1,\infty}^k \nonumber \\
	  & = \frac{1}{1- \| \bm{\mathcal C}( {\bm \Gamma}_{S,S}^{-\ast} ) \|_{1,\infty}
	                \| \bm{\mathcal C}( ({\bm \Delta}_\Gamma)_{S,S} ) \|_{1,\infty} }  \nonumber \\
	  & = \frac{1}{1- \kappa_\Gamma \| \bm{\mathcal C}( ({\bm \Delta}_\Gamma)_{S,S} ) \|_{1,\infty} }
			\overset{(\ref{aeqn429})}{<} \frac{1}{1-(1/3)} = \frac{3}{2} \, . \label{aeqn437}
\end{align}
Using (\ref{aeqn428}), (\ref{aeqn429}), (\ref{aeqn435}) and (\ref{aeqn437}), we have
\begin{align}
&	\| \bm{\mathcal C}( R({\bm \Delta}_\Gamma)) \|_\infty \; \le  \; 
	\frac{3}{2} \kappa_\Gamma^3 s_n  \| \bm{\mathcal C}( ({\bm \Delta}_\Gamma)_{S,S} ) \|_{\infty}^2 \nonumber \\
	  & \quad\quad <  \frac{3}{2} \kappa_\Gamma^3 s_n ( \epsilon^2 + 2M \epsilon)^2 \, . \label{aeqn439}
\end{align}
This proves (\ref{aeqn407}), from which (\ref{aeqn408}) immediately follows.

Using (\ref{eqn330}), (\ref{aeqn406}), (\ref{aeqn423}), (\ref{aeqn425}) and (\ref{aeqn428}) we have
\begin{align}
& \| \bm{\mathcal C}( \hat{\bm \Gamma}_{S,S}^{-1} - {\bm \Gamma}_{S,S}^{-\ast} ) \|_\infty
  \le \| \bm{\mathcal C}( R({\bm \Delta}_\Gamma)) \|_\infty \nonumber \\
	  & \quad\quad + 
	 \| \bm{\mathcal C}( {\bm \Gamma}_{S,S}^{-\ast} ({\bm \Delta}_\Gamma)_{S,S} {\bm \Gamma}_{S,S}^{-\ast} ) \|_\infty \nonumber \\
& \le \| \bm{\mathcal C}( R({\bm \Delta}_\Gamma)) \|_\infty + \| \bm{\mathcal C}( {\bm \Gamma}_{S,S}^{-\ast} )\|_{1,\infty}^2
       \|  \bm{\mathcal C}(  ({\bm \Delta}_\Gamma)_{S,S}   )  \|_\infty \nonumber \\
& < \kappa_\Gamma^2 (\epsilon^2+2M \epsilon) \big(1+ 1.5 s_n \kappa_\Gamma (\epsilon^2+2M \epsilon) \big)
  \, . \label{aeqn440}
\end{align}
This proves (\ref{aeqn409}). The claim (\ref{aeqn410}) follows by noting that $|S|=s_n$. This completes the proof. $\quad \Box$

{\bf Lemma 4}.  Assume (\ref{aeqn405}) and the following conditions:
\begin{align}
  & 0 < \alpha < 1  \mbox{ where } \alpha \mbox{ is as in } (\ref{eqn335}) \, ,
    \label{aeqn500}	\\
	& \epsilon  < \min \left\{ M, \frac{ \alpha \lambda_n }{2(2-\alpha)} \right\} \, , \label{aeqn502}	\\
	& \alpha C_\alpha \min\{\lambda_n,1\}  \ge  3 s_n \epsilon M \kappa_\Gamma B_s  \label{aeqn503}
\end{align}
where
\begin{align}
	C_\alpha & =  \frac{\alpha \lambda_n + 2 \epsilon \alpha - 4 \epsilon}
	          {2 M \alpha \lambda_n + \alpha \lambda_n + 2 \epsilon \alpha} \, , \label{aeqn504}	\\
	B_s & = \Big[ 1+ \kappa_\Gamma \Big( 3 s_n \epsilon M + \min\{s_n M^2,M_\Sigma^2\} \Big) \nonumber \\
	  & \quad\quad \times \big( 4.5 s_n \epsilon M \kappa_\Gamma +1 \big) \Big] \, . \label{aeqn505}
\end{align}
Then we have
\begin{itemize}
\item[(i)] $\mbox{bvec}(\hat{\bm \Delta}_{S^c}) = {\bm 0}$.
\item[(ii)]  $\| \bm{\mathcal C}(\hat{\bm \Delta} - {\bm \Delta}^\ast) \|_\infty 
  \le 2 \lambda_n \kappa_\Gamma + 3  s_n \epsilon M \kappa_\Gamma^2 
	  \big( 4.5 s_n \epsilon M \kappa_\Gamma +1 \big) \big( 2M + 2 \lambda_n \big) $ $\quad \bullet$
\end{itemize}
{\it Proof}. To establish part (i), we need to show that (\ref{aeqn261}) is true. Let $d$ denote the left-side of (\ref{aeqn261}). It follow from (\ref{aeqn260}) that
\begin{align} 
  \mbox{bvec}(\tilde{\bm \Delta}_S ) = & \hat{\bm \Gamma}_{S,S}^{-1} 
	  \Big( \mbox{bvec}((\hat{\bm \Sigma}_x-\hat{\bm \Sigma}_y)_S) - \lambda \, \mbox{bvec}({\bm Z}(\tilde{\bm \Delta}_S))\Big)
		 \, . \label{aeqn510}  
\end{align}
Substitute (\ref{aeqn510}) in the left-side of (\ref{aeqn261}) to yield
\begin{align} 
  d = & \| \hat{\bm \Gamma}_{e,S} \big[ \hat{\bm \Gamma}_{S,S}^{-1} 
	  \big( \mbox{bvec}((\hat{\bm \Sigma}_x-\hat{\bm \Sigma}_y)_S) 
		 - \lambda \, \mbox{bvec}({\bm Z}(\tilde{\bm \Delta}_S))\big) \big] \nonumber \\
		& \quad - \mbox{bvec}((\hat{\bm \Sigma}_x-\hat{\bm \Sigma}_y)_e) \|_2 \, . \label{aeqn520}  
\end{align}
At the true values we have
\begin{align} 
  {\bm 0} = & \frac{\partial L_\lambda({\bm \Delta}, {\bm \Sigma}_x^\ast , {\bm \Sigma}_y^\ast)}{\partial {\bm \Delta}} 
	   \Big|_{ {\bm \Delta} = {\bm \Delta}^\ast}
	     = {\bm \Sigma}_x^\ast {\bm \Delta}^\ast {\bm \Sigma}_y^\ast - ({\bm \Sigma}_x^\ast-{\bm \Sigma}_y^\ast) 
  \nonumber  
\end{align}
implying
\begin{align} 
   & {\bm \Gamma}^\ast \mbox{bvec}({\bm \Delta}^\ast )  - \mbox{bvec}({\bm \Sigma}_x^\ast-{\bm \Sigma}_y^\ast)  = {\bm 0} \, ,
  \label{aeqn530}  
\end{align}
which, noting that $({\bm \Delta}^\ast )_{S^c} = {\bm 0}$, can be rewritten as (cf.\ (\ref{aeqn227}))
\begin{align} 
  {\bm \Gamma}_{S,S}^\ast  \mbox{bvec}({\bm \Delta}_S^\ast ) = & \mbox{bvec}({\bm \Sigma}_x^\ast)_S 
	            - \mbox{bvec}({\bm \Sigma}_y^\ast)_S \, , \label{aeqn532}  \\
	{\bm \Gamma}_{e,S}^\ast  \mbox{bvec}({\bm \Delta}_S^\ast ) = & \mbox{bvec}({\bm \Sigma}_x^\ast)_e 
	            - \mbox{bvec}({\bm \Sigma}_y^\ast)_e \, . \label{aeqn534}  
\end{align}
Therefore, (${\bm A}^{-\ast} = ({\bm A}^\ast)^{-1}$),
\begin{align} 
	{\bm \Gamma}_{e,S}^\ast & {\bm \Gamma}_{S,S}^{-\ast} \big( \mbox{bvec}({\bm \Sigma}_x^\ast)_S 
	    - \mbox{bvec}({\bm \Sigma}_y^\ast)_S \big) \nonumber \\
		&	 \quad\quad -\mbox{bvec}({\bm \Sigma}_x^\ast)_e  + \mbox{bvec}({\bm \Sigma}_y^\ast)_e
							= {\bm 0}\, . \label{aeqn535}  
\end{align}
Recalling (\ref{aeqn400}) and using (\ref{aeqn535}) in (\ref{aeqn520}), 
\begin{align} 
  d = & \| \hat{\bm \Gamma}_{e,S}  \hat{\bm \Gamma}_{S,S}^{-1}  \mbox{bvec}(({\bm \Delta}_\Sigma)_S) \nonumber \\
		& + \big( \hat{\bm \Gamma}_{e,S} \hat{\bm \Gamma}_{S,S}^{-1} - {\bm \Gamma}_{e,S}^\ast {\bm \Gamma}_{S,S}^{-\ast} \big) 
		\big( \mbox{bvec}({\bm \Sigma}_x^\ast)_S - \mbox{bvec}({\bm \Sigma}_y^\ast)_S \big) \nonumber \\
		&- \lambda \, \hat{\bm \Gamma}_{e,S} \hat{\bm \Gamma}_{S,S}^{-1} 
		       \mbox{bvec}({\bm Z}(\tilde{\bm \Delta}_S))   - \mbox{bvec}(({\bm \Delta}_\Sigma)_e) \|_2 \, . \label{aeqn540}  
\end{align}

We now bound various terms in (\ref{aeqn540}). Note that $\hat{\bm \Gamma}_{e,S} \in \mathbb{R}^{m^2 \times (m^2s_n)}$, $\hat{\bm \Gamma}_{S,S}^{-1} \in \mathbb{R}^{(m^2s_n) \times (m^2s_n)}$, and $\mbox{bvec}(({\bm \Delta}_\Sigma)_S) \in \mathbb{R}^{m^2s_n}$ where ${\bm \Delta}_\Sigma \in \mathbb{R}^{(mp_n) \times (mp_n)}$. Consider ${\bm A}_{e,S} \in \mathbb{R}^{m^2 \times (m^2s_n)}$. Then 
\begin{align} 
	 \| {\bm A}_{e,S} \mbox{bvec}(({\bm \Delta}_\Sigma)_S)\|_2 & =
	  \| \sum_{f \in S} {\bm A}_{e,f} \mbox{vec}(({\bm \Delta}_\Sigma)_f)\|_2 \label{aeqn545}  
\end{align}
where edge $f \in S$, ${\bm A}_{e,f} \in \mathbb{R}^{m^2 \times m^2}$ and $({\bm \Delta}_\Sigma)_f \in \mathbb{R}^{m \times m}$. By the triangle inequality
\begin{align} 
	 \| {\bm A}_{e,S} \mbox{bvec}(({\bm \Delta}_\Sigma)_S)\|_2 & \le
	   \sum_{f \in S} \| {\bm A}_{e,f} \mbox{vec}(({\bm \Delta}_\Sigma)_f) \|_2 \, . \label{aeqn550}  
\end{align}
With ${\bm B}_{i.}$ denoting the $i$th row of matrix ${\bm B}$ and using Cauchy-Schwartz inequality, we have 
\begin{align} 
	& \| {\bm A}_{e,f} \mbox{vec}(({\bm \Delta}_\Sigma)_f)\|_2  = 
	   \Big( \sum_{i=1}^{m^2} \big( [{\bm A}_{e,f}]_{i.} \mbox{vec}(({\bm \Delta}_\Sigma)_f) \big)^2 \Big)^{1/2}  \nonumber \\
	& \quad \le \Big( \sum_{i=1}^{m^2} \| [{\bm A}_{e,f}]_{i.}\|_2^2 \, \|\mbox{vec}(({\bm \Delta}_\Sigma)_f)\|_2^2 \Big)^{1/2}  \nonumber \\
	& \quad = \|\mbox{vec}(({\bm \Delta}_\Sigma)_f)\|_2 \Big( \sum_{i=1}^{m^2} \| [{\bm A}_{e,f}]_{i.}\|_2^2  \Big)^{1/2}  \nonumber \\
	& \quad = \|\mbox{vec}(({\bm \Delta}_\Sigma)_f)\|_2 \, \| {\bm A}_{e,f}\|_F 	\, . \label{aeqn551}  
\end{align}
Therefore, using (\ref{aeqn550}) and (\ref{aeqn551}), 
\begin{align} 
	& \| {\bm A}_{e,S} \mbox{bvec}(({\bm \Delta}_\Sigma)_S)\|_2  \le
	   \Big(\sum_{f \in S} \| {\bm A}_{e,f} \|_F \Big) 
		 \max_{g \in S} \|\mbox{vec}(({\bm \Delta}_\Sigma)_g) \|_2 \nonumber \\
	& \quad\quad\quad  = \|\bm{\mathcal C}( {\bm A}_{e,S} ) \|_1 \, \|\bm{\mathcal C}( {\bm \Delta}_\Sigma ) \|_\infty \, . \label{aeqn552}  
\end{align} 
Using (\ref{aeqn552}), we have
\begin{align}
& \| \hat{\bm \Gamma}_{e,S}  \hat{\bm \Gamma}_{S,S}^{-1}  \mbox{bvec}(({\bm \Delta}_\Sigma)_S) \|_2 \nonumber \\
	& \quad
 \le \|\bm{\mathcal C}( \hat{\bm \Gamma}_{e,S}  \hat{\bm \Gamma}_{S,S}^{-1} ) \|_1 \, 
      \|\bm{\mathcal C}({\bm \Delta}_\Sigma) \|_\infty \, , \label{aeqn554} \\
	& \| \big( \hat{\bm \Gamma}_{e,S} \hat{\bm \Gamma}_{S,S}^{-1} - {\bm \Gamma}_{e,S}^\ast {\bm \Gamma}_{S,S}^{-\ast} \big) 
		\big( \mbox{bvec}({\bm \Sigma}_x^\ast)_S - \mbox{bvec}({\bm \Sigma}_y^\ast)_S \big) \|_2 \nonumber \\
	& \quad  \le \|\bm{\mathcal C}( \hat{\bm \Gamma}_{e,S} \hat{\bm \Gamma}_{S,S}^{-1}
	    - {\bm \Gamma}_{e,S}^\ast {\bm \Gamma}_{S,S}^{-\ast} ) \|_1 \, 
			 \|\bm{\mathcal C}({\bm \Sigma}_x^\ast- {\bm \Sigma}_y^\ast) \|_\infty \, , \label{aeqn555} \\
& \|  \hat{\bm \Gamma}_{e,S} \hat{\bm \Gamma}_{S,S}^{-1}  \mbox{bvec}({\bm Z}(\tilde{\bm \Delta}_S)) \|_2 \nonumber \\
	& \quad
 \le \|\bm{\mathcal C}( \hat{\bm \Gamma}_{e,S} \hat{\bm \Gamma}_{S,S}^{-1} ) \|_1 \, 
      \|\bm{\mathcal C}({\bm Z}(\tilde{\bm \Delta}_S)) \|_\infty \, , \label{aeqn556}	\\
& \| \mbox{bvec}(({\bm \Delta}_\Sigma)_e) \|_2 
 \le \|\bm{\mathcal C}({\bm \Delta}_\Sigma) \|_\infty \, . \label{aeqn557}	
\end{align}
By (\ref{aeqn215}), (\ref{aeqn400}) and (\ref{aeqn403})
\begin{align} 
	\|\bm{\mathcal C}({\bm \Sigma}_x^\ast- {\bm \Sigma}_y^\ast) \|_\infty & \le
	  \|\bm{\mathcal C}({\bm \Sigma}_x^\ast)\|_\infty + \|\bm{\mathcal C}({\bm \Sigma}_y^\ast)\|_\infty
		\le 2M \, , \label{aeqn560}  \\
	\|\bm{\mathcal C}({\bm Z}(\tilde{\bm \Delta}_S)) \|_\infty & \le 1 \, , \label{aeqn561}  \\
\|\bm{\mathcal C}({\bm \Delta}_\Sigma) \|_\infty & \le
	  \|\bm{\mathcal C}({\bm \Delta}_x)\|_\infty + \|\bm{\mathcal C}({\bm \Delta}_y)\|_\infty
		< 2 \epsilon \, . \label{aeqn562}
\end{align}
Using (\ref{aeqn540}) and (\ref{aeqn554})-\ref{aeqn562}),
\begin{align} 
  d \; < \; & 2 \epsilon \|\bm{\mathcal C}( \hat{\bm \Gamma}_{e,S}  \hat{\bm \Gamma}_{S,S}^{-1} ) \|_1  
	 + 2 M \| \bm{\mathcal C}( \hat{\bm \Gamma}_{e,S} \hat{\bm \Gamma}_{S,S}^{-1}
	    - {\bm \Gamma}_{e,S}^\ast {\bm \Gamma}_{S,S}^{-\ast} ) \|_1 \nonumber \\
		& + \lambda \, \|\bm{\mathcal C}( \hat{\bm \Gamma}_{e,S} \hat{\bm \Gamma}_{S,S}^{-1} ) \|_1 
		       + 2 \epsilon \, . \label{aeqn564}  
\end{align}
Therefore, $d < \lambda$ for any edge $e \in S^c$ if
\begin{align} 
  & U_{b1} := \max_{e \in S^c} 2 M \| \bm{\mathcal C}( \hat{\bm \Gamma}_{e,S} \hat{\bm \Gamma}_{S,S}^{-1}
	    - {\bm \Gamma}_{e,S}^\ast {\bm \Gamma}_{S,S}^{-\ast} ) \|_1 \nonumber \\
	& \quad\quad\quad + 2 \epsilon (1+\|\bm{\mathcal C}( \hat{\bm \Gamma}_{e,S}  \hat{\bm \Gamma}_{S,S}^{-1} ) \|_1)  
	  \le \alpha \lambda_n (1-C_\alpha) \, , \label{aeqn566}  \\
	& U_{b2} :=\max_{e \in S^c} \|\bm{\mathcal C}( \hat{\bm \Gamma}_{e,S} \hat{\bm \Gamma}_{S,S}^{-1} ) \|_1  
	 \le 1- (1-C_\alpha) \alpha \, . \label{aeqn567}
\end{align}

It remains to show that (\ref{aeqn566}) and (\ref{aeqn567}) are true under the assumptions of Lemma 4. Since
\begin{align}
	& \hat{\bm \Gamma}_{e,S} \hat{\bm \Gamma}_{S,S}^{-1} - {\bm \Gamma}_{e,S}^\ast {\bm \Gamma}_{S,S}^{-\ast}
	= (\hat{\bm \Gamma}_{e,S} - {\bm \Gamma}_{e,S}^\ast) {\bm \Gamma}_{S,S}^{-\ast} \nonumber \\
	&  +
	  {\bm \Gamma}_{e,S}^\ast (\hat{\bm \Gamma}_{S,S}^{-1} -  {\bm \Gamma}_{S,S}^{-\ast})
		+ (\hat{\bm \Gamma}_{e,S} - {\bm \Gamma}_{e,S}^\ast) (\hat{\bm \Gamma}_{S,S}^{-1} -  {\bm \Gamma}_{S,S}^{-\ast})
		\, ,  \label{aeqn570}	
\end{align}
we have
\begin{align}
	& \|\bm{\mathcal C}(\hat{\bm \Gamma}_{e,S} \hat{\bm \Gamma}_{S,S}^{-1} 
	 - {\bm \Gamma}_{e,S}^\ast {\bm \Gamma}_{S,S}^{-\ast} ) \|_\infty \nonumber \\
	& \quad
	\le \|\bm{\mathcal C}(\hat{\bm \Gamma}_{e,S} - {\bm \Gamma}_{e,S}^\ast) \|_\infty
	 \|\bm{\mathcal C}( {\bm \Gamma}_{S,S}^{-\ast} ) \|_{1,\infty} \nonumber \\
	& \quad\quad +
	 \|\bm{\mathcal C}( {\bm \Gamma}_{e,S}^\ast ) \|_\infty
	  \|\bm{\mathcal C}(\hat{\bm \Gamma}_{S,S}^{-1} -  {\bm \Gamma}_{S,S}^{-\ast}) \|_{1,\infty} \nonumber \\
	& \quad\quad
		+ \|\bm{\mathcal C}(\hat{\bm \Gamma}_{e,S} - {\bm \Gamma}_{e,S}^\ast) \|_\infty 
		\|\bm{\mathcal C}(\hat{\bm \Gamma}_{S,S}^{-1} -  {\bm \Gamma}_{S,S}^{-\ast}) \|_{1,\infty}
		\, . \label{aeqn572}	
\end{align}
With edge $e=\{i,k\} \in S^c$ and edge $f=\{j,\ell\} \in S$, consider
\begin{align}
	& \hat{\bm \Gamma}_{e,f} - {\bm \Gamma}_{e,f}^\ast 
	         =  \hat{\bm \Gamma}_{ik,j \ell} - {\bm \Gamma}_{ik,j \ell}^\ast  \nonumber \\
	& 
	  = \hat{\bm \Sigma}_y^{(ij)} \otimes \hat{\bm \Sigma}_x^{(k \ell)}
		   - {\bm \Sigma}_y^{\ast(ij)} \otimes {\bm \Sigma}_x^{\ast(k \ell)}  \nonumber \\
	& = {\bm \Delta}_y^{(ij)} \otimes {\bm \Delta}_x^{(k \ell)} +
	      {\bm \Sigma}_y^{\ast(ij)} \otimes  {\bm \Delta}_x^{(k \ell)} + {\bm \Delta}_y^{(ij)} \otimes {\bm \Sigma}_x^{\ast(k \ell)}
		 .  \label{aeqn574}	
\end{align}
It then follows that
\begin{align}
	& |\bm{\mathcal C}(\hat{\bm \Gamma}_{e,f} - {\bm \Gamma}_{e,f}^\ast )|					
		\le  \| {\bm \Delta}_y^{(ij)}\|_F \, \| {\bm \Delta}_x^{(k \ell)} \|_F \nonumber \\
	& \quad +
	     \| {\bm \Sigma}_y^{\ast(ij)} \|_F \, \| {\bm \Delta}_x^{(k \ell)} \|_F
			+ \| {\bm \Delta}_y^{(ij)} \|_F \, \| {\bm \Sigma}_x^{\ast(k \ell)} \|_F \nonumber \\
	& \le \epsilon_y \epsilon_x + M \epsilon_x + M \epsilon_y < \epsilon^2 + 2M \epsilon
		 .  \label{aeqn575}	
\end{align}
Hence
\begin{align}
	\|\bm{\mathcal C}(\hat{\bm \Gamma}_{e,S} - {\bm \Gamma}_{e,S}^\ast) \|_\infty < \; & \epsilon^2 + 2M \epsilon \, ,
	  \label{aeqn576} \\
  \|\bm{\mathcal C}(\hat{\bm \Gamma}_{e,S} - {\bm \Gamma}_{e,S}^\ast) \|_1 < \; & s_n (\epsilon^2 + 2M \epsilon)
		 .  \label{aeqn577}	
\end{align}
Since ${\bm \Gamma}_{e,f}^\ast = {\bm \Sigma}_y^{\ast(ij)} \otimes {\bm \Sigma}_x^{\ast(k \ell)}$, we have $|{\bm \Gamma}_{e,f}^\ast | \le M^2$ and 
\begin{align}
	 \|\bm{\mathcal C}({\bm \Gamma}_{e,S}^\ast) \|_\infty & \le M^2 , \quad 
	 \|\bm{\mathcal C}({\bm \Gamma}_{e,S}^\ast) \|_1 \le s_n M^2 \,
		 .  \label{aeqn578}	
\end{align}
Alternatively, with $e=\{i,k\} \in S^c$ and $f=\{j,\ell\} \in S$,
\begin{align}
	& \|\bm{\mathcal C}({\bm \Gamma}_{e,S}^\ast) \|_1 
	 = \sum_{f \in S} |\bm{\mathcal C}( {\bm \Sigma}_y^{\ast(ij)} \otimes {\bm \Sigma}_x^{\ast(k \ell)} )| \nonumber \\
	&  \quad \le \sum_{\{j,\ell\} \in S} \| {\bm \Sigma}_y^{\ast(ij)}\|_F \, \| {\bm \Sigma}_x^{\ast(k \ell)} \|_F \nonumber \\
	&  \quad \le ( \sum_{j=1}^p \| {\bm \Sigma}_y^{\ast(ij)}\|_F ) ( \sum_{\ell =1}^p \| {\bm \Sigma}_x^{\ast(k \ell)} \|_F ) \nonumber \\
	&  \quad \le \|\bm{\mathcal C}( {\bm \Sigma}_y^{\ast} ) \|_{1,\infty} \, \|\bm{\mathcal C}( {\bm \Sigma}_x^{\ast} ) \|_{1,\infty}
	  \le M_\Sigma^2
		 \, .  \label{aeqn579}	
\end{align}
From (\ref{aeqn570})-(\ref{aeqn579}) and Lemma 3, we have 
\begin{align}
	& \|\bm{\mathcal C}(\hat{\bm \Gamma}_{e,S} \hat{\bm \Gamma}_{S,S}^{-1} 
	 - {\bm \Gamma}_{e,S}^\ast {\bm \Gamma}_{S,S}^{-\ast} ) \|_1 \nonumber \\
	& \quad
	\le \|\bm{\mathcal C}(\hat{\bm \Gamma}_{e,S} - {\bm \Gamma}_{e,S}^\ast) \|_1
	 \|\bm{\mathcal C}( {\bm \Gamma}_{S,S}^{-\ast} ) \|_{1,\infty} \nonumber \\
	& \quad\quad +
	 \|\bm{\mathcal C}( {\bm \Gamma}_{e,S}^\ast ) \|_1
	  \|\bm{\mathcal C}(\hat{\bm \Gamma}_{S,S}^{-1} -  {\bm \Gamma}_{S,S}^{-\ast}) \|_{1,\infty} \nonumber \\
	& \quad\quad
		+ \|\bm{\mathcal C}(\hat{\bm \Gamma}_{e,S} - {\bm \Gamma}_{e,S}^\ast) \|_1 
		\|\bm{\mathcal C}(\hat{\bm \Gamma}_{S,S}^{-1} -  {\bm \Gamma}_{S,S}^{-\ast}) \|_{1,\infty} \nonumber \\
	& \quad
	\le s_n (\epsilon^2 + 2M \epsilon) \kappa_\Gamma + \big[ \min\{s_nM^2, M_\Sigma^2 \} 
	  + s_n (\epsilon^2 + 2M \epsilon) \big] \nonumber \\
	& \quad\quad\quad \times \big[s_n (\epsilon^2 + 2M \epsilon) \kappa_\Gamma^2 \big] 
	      \big(1+1.5 s_n (\epsilon^2 + 2M \epsilon) \kappa_\Gamma \big) \nonumber \\
  & \quad \overset{\epsilon < M}{\le} 3 s_n \epsilon M \kappa_\Gamma B_s
	  \, \le \, \alpha C_\alpha \min\{\lambda_n,1\}
		\, , \label{aeqn580}	
\end{align}
where $B_s$ is as in (\ref{aeqn505}) and we used $\epsilon < M$ to infer $\epsilon^2 + 2M \epsilon < 3M \epsilon$.
Using the triangle inequality $|a|-|b| \le |a-b| \le |a| + |b|$, we have 
$\|\bm{\mathcal C}(\hat{\bm \Gamma}_{e,S} \hat{\bm \Gamma}_{S,S}^{-1} - {\bm \Gamma}_{e,S}^\ast {\bm \Gamma}_{S,S}^{-\ast} ) \|_1$
	 $\ge$ $\|\bm{\mathcal C}(\hat{\bm \Gamma}_{e,S} \hat{\bm \Gamma}_{S,S}^{-1}) \|_1$
	    $- \|\bm{\mathcal C}({\bm \Gamma}_{e,S}^\ast {\bm \Gamma}_{S,S}^{-\ast} ) \|_1$ ,
which, using  (\ref{aeqn500}) and (\ref{aeqn580}), leads to
\begin{align}
	& \|\bm{\mathcal C}(\hat{\bm \Gamma}_{e,S} \hat{\bm \Gamma}_{S,S}^{-1}) \|_1
		\le \|\bm{\mathcal C}({\bm \Gamma}_{e,S}^\ast {\bm \Gamma}_{S,S}^{-\ast} ) \|_1 
		 + \alpha C_\alpha \min\{\lambda_n,1\}  \nonumber \\
  & \quad \le 1- \alpha + \alpha C_\alpha \min\{\lambda_n,1\} \le 1- (1-C_\alpha) \alpha
		 \, .  \label{aeqn581}	
\end{align}
This establishes (\ref{aeqn567}). To show (\ref{aeqn566}), using (\ref{aeqn580})-(\ref{aeqn581}),
\begin{align}
	 U_{b1} \; \le \; &  2M \alpha C_\alpha \min\{\lambda_n,1\} + 2 \epsilon(1+ 1- (1-C_\alpha) \alpha)
	   \nonumber \\
   \le \; & 2M \alpha C_\alpha \lambda_n + 2 \epsilon(2- (1-C_\alpha) \alpha)
	  \overset{(\ref{aeqn504})}{=} \alpha \lambda_n (1-C_\alpha)
		 \, .  \label{aeqn584}	
\end{align}
This proves (\ref{aeqn566}), and thus, part (i) of Lemma 4.

We now turn to the proof of Lemma 4(ii). Since $\hat{\bm \Delta} = \tilde{\bm \Delta}$, for any edge $\{k, \ell \} \in S$, we have 
\begin{align}
	& \| (\hat{\bm \Delta} - {\bm \Delta}^\ast )^{(k \ell)} \|_F 
	   = \| (\tilde{\bm \Delta} - {\bm \Delta}^\ast )^{(k \ell)} \|_F \nonumber \\
  & \quad
	  = \| \mbox{vec}(\tilde{\bm \Delta}^{(k \ell)}) - \mbox{vec}(({\bm \Delta}^{\ast})^{(k \ell)}) \|_2
		 \, .  \label{aeqn590}	
\end{align}
Using (\ref{aeqn260}) and (\ref{aeqn532})
\begin{align}
	& \mbox{bvec}( (\tilde{\bm \Delta} - {\bm \Delta}^\ast )_S ) 
	= \hat{\bm \Gamma}_{S,S}^{-1} \mbox{bvec}(({\bm \Delta}_\Sigma)_S) 
	  + ( \hat{\bm \Gamma}_{S,S}^{-1} - {\bm \Gamma}_{S,S}^{-\ast} )\nonumber \\
  & \quad \times 
	 \mbox{bvec}(({\bm \Sigma}_x^\ast - {\bm \Sigma}_y^\ast)_S) 
	  - \lambda_n \hat{\bm \Gamma}_{S,S}^{-1} \, \mbox{bvec}({\bm Z}(\tilde{\bm \Delta}_S))
		 \, .  \label{aeqn592}	
\end{align}
Since $ \hat{\bm \Gamma}_{S,S}^{-1} = \hat{\bm \Gamma}_{S,S}^{-1} - {\bm \Gamma}_{S,S}^{-\ast} +{\bm \Gamma}_{S,S}^{-\ast} \, ,$ 
\begin{align}
	& \|\bm{\mathcal C}( \hat{\bm \Gamma}_{S,S}^{-1} ) \|_{1,\infty} \le 
	  \|\bm{\mathcal C}( \hat{\bm \Gamma}_{S,S}^{-1} - {\bm \Gamma}_{S,S}^{-\ast}  ) \|_{1,\infty} +
	\|\bm{\mathcal C}( {\bm \Gamma}_{S,S}^{-\ast} ) \|_{1,\infty}
		 \, .  \label{aeqn594}	
\end{align}
By (\ref{aeqn592}), for any edge $f = \{k, \ell\} \in S$, we have
\begin{align}
	& \| \mbox{vec}( (\tilde{\bm \Delta} - {\bm \Delta}^\ast)^{(k \ell)}  ) \|_2
	\le \| (\hat{\bm \Gamma}_{S,S}^{-1} - {\bm \Gamma}_{S,S}^{-\ast})_{f,S}  \nonumber \\
  & \quad\quad\quad \times
	  \mbox{bvec} \big( ({\bm \Delta}_\Sigma)_S + ({\bm \Sigma}_x^\ast - {\bm \Sigma}_y^\ast)_S
		- \lambda_n {\bm Z}(\tilde{\bm \Delta}_S) \big) \|_2  \nonumber \\
  & \quad + \| ({\bm \Gamma}_{S,S}^{-\ast})_{f,S} \,
	 \mbox{bvec} \Big( ({\bm \Delta}_\Sigma)_S  
		- \lambda_n {\bm Z}(\tilde{\bm \Delta}_S) \Big) \|_2  \nonumber \\
  &  \le \|\bm{\mathcal C}( \hat{\bm \Gamma}_{S,S}^{-1} - {\bm \Gamma}_{S,S}^{-\ast}  ) \|_{1,\infty} \,
	  \Big( \|\bm{\mathcal C}( {\bm \Delta}_\Sigma ) \|_{\infty} 
		 + \|\bm{\mathcal C}( {\bm \Sigma}_x^\ast - {\bm \Sigma}_y^\ast ) \|_{\infty}   \nonumber \\
  & \quad\quad + \lambda_n \Big) + \|\bm{\mathcal C}( {\bm \Gamma}_{S,S}^{-\ast} ) \|_{1,\infty} \,
	   \big( \|\bm{\mathcal C}( {\bm \Delta}_\Sigma ) \|_{\infty} + \lambda_n \big)  \nonumber \\
  &  \le s_n \kappa_\Gamma^2 (\epsilon^2 + 2M \epsilon)  
	      \big(1+1.5 s_n (\epsilon^2 + 2M \epsilon) \kappa_\Gamma \big)  \nonumber \\
  &  \quad\quad\quad \times (2 \epsilon + 2 M + \lambda_n) + \kappa_\Gamma (2 \epsilon + \lambda_n)  =: U_{b3}
		 \, .  \label{aeqn596}	
\end{align}
By (\ref{aeqn502}), for $0 < \alpha < 1$, we have $2 \epsilon < \alpha \lambda_n /(2-\alpha) < \alpha \lambda_n < \lambda_n$. Therefore, $\kappa_\Gamma (2 \epsilon + \lambda_n) < 2 \kappa_\Gamma \lambda_n$ and $2 \epsilon + 2 M + \lambda_n < 2M + 2 \lambda_n$. Since $\epsilon < M$ by (\ref{aeqn502}), we also have $\epsilon^2 + 2M \epsilon < 3M \epsilon$. Using these relations and (\ref{aeqn596}), it follows that
\begin{align*}
	U_{b3} \; \le \; &  3  s_n \epsilon M \kappa_\Gamma^2 
	  \big( 1+ 4.5 s_n \epsilon M \kappa_\Gamma  \big) \big( 2M + 2 \lambda_n \big) + 2 \lambda_n \kappa_\Gamma 
		 \, .  
\end{align*}
Finally, 
\begin{align*}
  \| \bm{\mathcal C}(\hat{\bm \Delta} - {\bm \Delta}^\ast) \|_\infty = & \max_{f = \{k, \ell\} \in S} \| \mbox{vec}( (\tilde{\bm \Delta} - {    \bm \Delta}^\ast)^{(k \ell)}  ) \|_2 \, ,
\end{align*}
proving the desired result. $\quad \Box$

{\it Proof of Theorem 1}. Here we first show that under the sufficient conditions of Theorem 1, the assumptions of Lemmas 2-4 holds true. 
We pick $\epsilon = C_0 \sqrt{ln(p_n)/n}$, implying, by Lemma 2, that $\|\bm{\mathcal C}( \hat{\bm \Sigma}_x - {\bm \Sigma}_x^\ast ) \|_{\infty} \le \epsilon$ and $\|\bm{\mathcal C}( \hat{\bm \Sigma}_y - {\bm \Sigma}_y^\ast ) \|_{\infty} \le \epsilon$. with probability $\ge 1-2/p_n^{\tau -2}$, $\tau > 2$. We first show that with this choice of $\epsilon$, condition (\ref{aeqn502}) of Lemma 4 holds true. By the choice of $\lambda_n$, we have $\lambda_n \ge 8 \epsilon/\alpha$. Clearly, for $0 < \alpha < 1$, $(\alpha/8) < \alpha/(4(2-\alpha))$. Therefore
\begin{align}
  \epsilon & \le \frac{\alpha \lambda_n}{8} < \frac{\alpha \lambda_n}{4(2-\alpha)} < \frac{\alpha \lambda_n}{2(2-\alpha)}
	  \, .  \label{aeqn600}
\end{align}
By a choice of $n$ in (\ref{eqn352}), we have $n > C_0^2 \ln(p_n)/M^2$. Hence, $\epsilon = C_0 \sqrt{ln(p_n)/n} < M$. Thus, (\ref{aeqn502}) of Lemma 4 holds true. Next we show that condition (\ref{aeqn405}) of Lemma 3 holds. By a choice of $n$ in (\ref{eqn352}), we have $n > 81 C_0^2 \ln(p_n) M^2 s_n^2 \kappa_\Gamma^2$. Therefore,
\begin{align}
  \kappa_\Gamma & < \frac{1}{C_0} \sqrt{\frac{n}{\ln (p_n)}} \times \frac{1}{9 s_n M}
	    = \frac{1}{9 s_n M \epsilon}  < \frac{1}{3 s_n (\epsilon^2 + 2 M \epsilon)}
	    \label{aeqn602}
\end{align}
since $\epsilon < M$. This proves (\ref{aeqn405}) of Lemma 3 holds.

Now we show that (\ref{aeqn503}) of Lemma 4 holds. Since $ \epsilon < \alpha \lambda_n / (4(2-\alpha))$, by (\ref{aeqn504}), we have
\begin{align}
	 C_\alpha & =  \frac{\alpha \lambda_n + 2 \epsilon \alpha - 4 \epsilon}
	          {2 M \alpha \lambda_n + \alpha \lambda_n + 2 \epsilon \alpha}
			> \frac{\alpha \lambda_n - 4 \epsilon }
	          { 2 M \alpha \lambda_n + \alpha \lambda_n + 2 \epsilon \alpha} \nonumber \\
  & > \frac{\alpha \lambda_n - \alpha \lambda_n / (2-\alpha) }
	          {\alpha \lambda_n (2M + 1) + 2 \epsilon \alpha } 
		= \frac{1-\alpha} {(2-\alpha) \big[ 2M + 1 + \frac{2 \epsilon \alpha }{\lambda_n \alpha} \big] }
							\nonumber \\
  &  > \frac{1-\alpha} {(2-\alpha) (2M + 1) + \alpha }  = \bar{C}_\alpha
		   \label{aeqn606}	
\end{align}
where in the last inequality above, we used  $\epsilon < \alpha \lambda_n / (2(2-\alpha))$ from (\ref{aeqn600}). 
Consider the right-side $3 s_n \epsilon M \kappa_\Gamma B_s$ of (\ref{aeqn503}). From (\ref{aeqn602}), $\kappa_\Gamma < 1/(9 s_n M \epsilon)$. Therefore,
\begin{align}
	B_s < &  1+ \kappa_\Gamma \big( \min\{s_n M^2,M_\Sigma^2\} + \frac{1}{3 \kappa_\Gamma} \big) \big( \frac{1}{2} +1 \big) \nonumber \\
    = & 1.5 + 1.5 \kappa_\Gamma \min\{s_n M^2,M_\Sigma^2\} = C_{M\kappa} \, .
		   \label{aeqn608}	
\end{align}
By (\ref{eqn350}), $\lambda_n \ge 4.5 \epsilon (\alpha \bar{C}_\alpha)^{-1} s_n M \kappa_\Gamma (1+ \kappa_\Gamma \min\{s_n M^2,M_\Sigma^2\})$. Hence, using (\ref{aeqn606}), we have
\begin{align}
	3 s_n \epsilon M \kappa_\Gamma B_s < &  \alpha \bar{C}_\alpha \lambda_n < \alpha {C}_\alpha \lambda_n  \, ,
		   \label{aeqn610}	
\end{align}
proving part of (\ref{aeqn503}) for our choice of $\lambda_n$. To show that we also have $3 s_n \epsilon M \kappa_\Gamma B_s < \alpha \bar{C}_\alpha$, consider the choice of $n$ in (\ref{eqn352}) given by
\begin{align}
	n > \; &  {C}_0^2 \ln(p_n) 	\, \frac{9 s_n^2}{(\alpha \bar{C}_\alpha)^2}  (\kappa_\Gamma M C_{M\kappa})^2 \, .
		   \label{aeqn612}	
\end{align} 
Then
\begin{align}
	\epsilon =\;  &  {C}_0  \sqrt{\frac{\ln(p_n)}{n}} \, 	
	  < \; \frac{\alpha \bar{C}_\alpha}{3 s_n M \kappa_\Gamma C_{M\kappa}} \, ,
		   \label{aeqn614}	
\end{align}
and from (\ref{aeqn610}),
\begin{align}
	3 s_n \epsilon M \kappa_\Gamma B_s < \; &  \alpha \bar{C}_\alpha  < \alpha {C}_\alpha   \, .
		   \label{aeqn616}	
\end{align}
Thus, all assumptions of Lemma 4 hold true. 

Therefore, Lemma 4(i) applies, proving Theorem 1(ii). By (\ref{aeqn602}), $9 s_n \epsilon M \kappa_\Gamma < 1$. Using this fact in Lemma 4(ii), 
\begin{align}
  & \| \bm{\mathcal C}(\hat{\bm \Delta} - {\bm \Delta}^\ast) \|_\infty 
  \le 2 \lambda_n \kappa_\Gamma + 9  s_n \epsilon M \kappa_\Gamma^2 \big( M +  \lambda_n \big) \nonumber \\
  & \quad\quad
	\le 3 \lambda_n \kappa_\Gamma + 9  s_n \epsilon M^2 \kappa_\Gamma^2 \, .
		   \label{aeqn618}
\end{align}
Since, by (\ref{eqn350}), $3 \lambda_n \kappa_\Gamma = {C}_0  \sqrt{\ln(p_n)/n} \, C_{b1}$ and we picked $\epsilon = {C}_0  \sqrt{\ln(p_n)/n}$, we have
\begin{align*}
  & \| \bm{\mathcal C}(\hat{\bm \Delta} - {\bm \Delta}^\ast) \|_\infty 
  \le (C_{b1} + C_{b2}) {C}_0 \sqrt{\ln(p_n)/n}
   \, ,
\end{align*}
proving Theorem 1(i). To prove part (iii), since $\hat{\bm \Delta}_{S^c} = \tilde{\bm \Delta}_{S^c} = {\bm \Delta}_{S^c}^\ast ={\bm 0}$, we have
\begin{align}
  & \| \bm{\mathcal C}(\hat{\bm \Delta} - {\bm \Delta}^\ast) \|_F
	= \Big( \sum_{ \{k , \ell \} \in S} \| \hat{\bm \Delta}^{(k \ell)} - ({\bm \Delta}^\ast)^{(k \ell)} \|_F^2 \Big)^{1/2} \nonumber \\
  & \quad\quad\quad \le \| \bm{\mathcal C}(\hat{\bm \Delta} - {\bm \Delta}^\ast) \|_\infty \, \sqrt{s_n}
   \, .
		   \label{aeqn622}
\end{align}
Finally, to establish part (iv), note that parts (i)-(iii) hold with probability $> 1-2/p_n^{\tau-2}$ (with high probability (w.h.p.)). Recall that ${\cal G}_\Delta = \left( V, {\cal E}_\Delta \right)$ denotes the MA differential graph with edgeset ${\cal E}_\Delta = \{ \{ k, \ell \} \,:\, \| {\bm \Delta}^{(k \ell)} \|_F > 0 \}$. Let ${\cal G}_{\Delta^\ast}$ and ${\cal G}_{\hat \Delta}$ denoted true and estimated graphs based on ${\bm \Delta}^\ast$ and $\hat{\bm \Delta}$, respectively. If $\min_{(k,\ell) \in S} \| ({\bm \Delta}^\ast)^{(k \ell)} \|_F \ge 2\| \bm{\mathcal C}(\hat{\bm \Delta} - {\bm \Delta}^\ast) \|_\infty$, then $\bm{\mathcal C}(\hat{\bm \Delta} - {\bm \Delta}^\ast) \|_\infty = \bm{\mathcal C}((\hat{\bm \Delta} - {\bm \Delta}^\ast)_S) \|_\infty \le (1/2) \min_{(k,\ell) \in S} \| ({\bm \Delta}^\ast)^{(k \ell)} \|_F$, therefore, $ \min_{(k,\ell) \in S} \| (\hat{\bm \Delta}_S)^{(k \ell)} \|_F \ge (1/2) \min_{(k,\ell) \in S} \| ({\bm \Delta}^\ast)^{(k \ell)} \|_F > 0$, while $\hat{\bm \Delta}_{S^c} ={\bm 0}$ w.h.p. $\quad \Box$

\section{Technical Lemmas and Proof of Theorem 2} \label{append2}
In order to invoke \cite{Negahban2012}, we first vectorize (\ref{eqn15}), using ${\bm \theta} = \mbox{bvec}({\bm \Delta}) \in \mathbb{R}^{m^2p^2}$, as (cf.\ (\ref{aeqn220}))
\begin{equation}
	{\cal L}({\bm \theta}) = \frac{1}{2}  {\bm \theta}^\top (\hat{\bm \Sigma}_y \boxtimes  \hat{\bm \Sigma}_x) {\bm \theta}
	  - {\bm \theta}^\top \mbox{bvec}(\hat{\bm \Sigma}_x-\hat{\bm \Sigma}_y) 
	\label{aeqn700}
\end{equation}
where previous $L({\bm \Delta}, \hat{\bm \Sigma}_x , \hat{\bm \Sigma}_y)$ is now ${\cal L}({\bm \theta})$. To include sparse-group penalty, recall that the submatrix ${\bm \Delta}^{(k \ell)}$ of ${\bm \Delta}$ corresponds to the edge $\{k,\ell\}$ of the MA graph. We denote its vectorized version as ${\bm \theta}_{Gt} \in \mathbb{R}^{m^2}$ (subscript $G$ for grouped variables \cite{Negahban2012}) with index $t=1,2, \cdots , p^2$. Then ${\bm \theta}_{Gt} = \mbox{vec}({\bm \Delta}^{(k \ell)})$ where $t=(k-1)p+\ell$, $\ell = t\mod{p}$, and $k=\lfloor t/p \rfloor +1$.  Using this notation, the penalty $\lambda \sum_{k, \ell=1}^p \| {\bm \Delta}^{(k \ell)} \|_F = \lambda \sum_{t=1}^{p^2} \| {\bm \theta}_{Gt} \|_2$. In the notation of \cite{Negahban2012}, the regularization penalty
\begin{equation}
	{\cal R}({\bm \theta}) = \|{\bm \theta}\|_{\bar{\cal G},2} := \sum_{t=1}^{p^2} \| {\bm \theta}_{Gt} \|_2 
	\label{aeqn705}
\end{equation}
where the index set $\{1,2, \cdots , (mp)^2\}$ is partitioned into a set of $N_G=p^2$ disjoint groups $\bar{\cal G} = \{G_1, G_2, \cdots , G_{p^2} \}$. As shown in \cite[Sec.\ 2.2]{Negahban2012}, ${\cal R}({\bm \theta})$ is a norm. The counterpart to penalized $L_\lambda({\bm \Delta}, \hat{\bm \Sigma}_x , \hat{\bm \Sigma}_y)$ of (\ref{eqn20}) is (we denote $\lambda$ by $\lambda_n$, as in Appendix \ref{append1})
\begin{align}
	{\cal L}_\lambda({\bm \theta}) = & {\cal L}({\bm \theta}) + \lambda_n {\cal R}({\bm \theta}) \, .
	\label{aeqn707}
\end{align}

As discussed in \cite[Sec.\ 2.2]{Negahban2012}, w.r.t.\ the usual Euclidean inner product $\langle {\bm u} , {\bm v} \rangle = {\bm u}^\top {\bm v}$ for ${\bm u} , {\bm v} \in \mathbb{R}^{m^2p^2}$ and given any subset $S_{\bar{\cal G}} \subseteq \{1,2, \cdots , N_G \}$ of group indices, define the subspace
\begin{equation}
	{\mathcal M} = \{ {\bm \theta} \in \mathbb{R}^{m^2p^2} \, | \, 
	      {\bm \theta}_{Gt} = {\bm 0} \mbox{ for all } t \not\in S_{\bar{\cal G}} \}
	\label{aeqn710}
\end{equation}
and its orthogonal complement
\begin{equation}
	{\mathcal M}^\perp = \{ {\bm \theta} \in \mathbb{R}^{m^2p^2} \, | \, 
	      {\bm \theta}_{Gt} = {\bm 0} \mbox{ for all } t \in S_{\bar{\cal G}} \} \, .
	\label{aeqn712}
\end{equation}
The chosen ${\cal R}({\bm \theta})$ is decomposable w.r.t.\ $({\mathcal M}, {\mathcal M}^\perp)$ since ${\cal R}({\bm \theta}^{(1)}+{\bm \theta}^{(2)}) = {\cal R}({\bm \theta}^{(1)}) + {\cal R}({\bm \theta}^{(2)})$ for any ${\bm \theta}^{(1)} \in {\mathcal M}$ and ${\bm \theta}^{(2)} \in {\mathcal M}^\perp$ \cite[Sec.\ 2.2, Example 2]{Negahban2012}. 

In order to invoke \cite{Negahban2012}, we need the dual norm ${\cal R}^\ast$ of regularizer ${\cal R}$ w.r.t.\ the inner product $\langle {\bm u} , {\bm v} \rangle = {\bm u}^\top {\bm v}$. It is given by \cite[Eqns.\ (14)-(15)]{Negahban2012}
\begin{align}
	{\cal R}^\ast({\bm v}) = & \sup_{{\cal R}({\bm u}) \le 1} \langle {\bm u} , {\bm v} \rangle \; = \;
	      \max_{t=1, 2, \cdots p^2} \| {\bm u}_{Gt} \|_2 \, .
	\label{aeqn714}
\end{align}
We also need the subspace compatibility index \cite{Negahban2012}, defined as
\begin{align}
	\Psi({\mathcal M})= & \sup_{{\bm u} \in {\mathcal M} \backslash \{0\}} \frac{{\cal R}({\bm u})}{\|{\bm u }\|_2}  \, .
	\label{aeqn716}
\end{align}
For group lasso penalty, $\Psi({\mathcal M}) = \sqrt{s_n} \,$ \cite[Sec.\ 9.2 (Supplementary)]{Negahban2012}, where $s_n = |S_{\bar{\cal G}}| =$ number of edges in the true MA differential graph. We need to establish a restricted strong  convexity condition \cite{Negahban2012} on ${\cal L}({\bm \theta})$. With ${\bm \theta}^\ast = \mbox{bvec}({\bm \Delta}^\ast)$ denoting the true value, and ${\bm \theta} = {\bm \theta}^\ast + \tilde{\bm \theta}$, define
\begin{align}
	\delta {\cal L}(\tilde{\bm \theta}, {\bm \theta}^\ast) = & {\cal L}({\bm \theta}^\ast+\tilde{\bm \theta}) -
	   {\cal L}({\bm \theta}^\ast) -\langle \nabla {\cal L}({\bm \theta}^\ast) , \tilde{\bm \theta} \rangle 
	\label{aeqn720}
\end{align}
where the gradient $\nabla {\cal L}({\bm \theta}^\ast)$ at ${\bm \theta} = {\bm \theta}^\ast$ is
\begin{align}
  \nabla {\cal L}({\bm \theta}^\ast) = & (\hat{\bm \Sigma}_y \boxtimes  \hat{\bm \Sigma}_x) {\bm \theta}^\ast 
	 - \mbox{bvec}(\hat{\bm \Sigma}_x-\hat{\bm \Sigma}_y) \, .
	\label{aeqn722}
\end{align}
Hence (\ref{aeqn720}) simplifies to
\begin{align}
	\delta {\cal L}(\tilde{\bm \theta}, {\bm \theta}^\ast) = & 
	   \tilde{\bm \theta}^\top (\hat{\bm \Sigma}_y \boxtimes  \hat{\bm \Sigma}_x) \tilde{\bm \theta}
		  = \tilde{\bm \theta}^\top  \hat{\bm \Gamma} \tilde{\bm \theta}  \, ,
	\label{aeqn724}
\end{align}
which may be rewritten as
\begin{align}
	\delta {\cal L}(\tilde{\bm \theta}, {\bm \theta}^\ast) = & \tilde{\bm \theta}^\top  {\bm \Gamma}^\ast \tilde{\bm \theta}
	   + \tilde{\bm \theta}^\top  (\hat{\bm \Gamma}-{\bm \Gamma}^\ast) \tilde{\bm \theta}  \, .
	\label{aeqn726}
\end{align}
By the sparsity assumption, ${\bm \theta}^\ast = {\bm \theta}^\ast_{\mathcal M}$, hence, ${\bm \theta}^\ast_{{\mathcal M}^\perp} = {\bm 0}$, where 
${\bm \theta}_{\mathcal M}$ and ${\bm \theta}_{{\mathcal M}^\perp}$ denote projection of ${\bm \theta}$ on subspaces ${\mathcal M}$ and ${\mathcal M}^\perp$, respectively.

Similar to (\ref{eqn25}), suppose
\begin{equation}
  \hat{\bm \theta} = \arg\min_{\bm \theta} \big\{{\cal L}({\bm \theta}) + \lambda_n {\cal R}({\bm \theta}) \big\} \, , \label{aeqn730}
\end{equation}
and we consider (\ref{aeqn720}) and (\ref{aeqn724}) with $\hat{\bm \theta} = {\bm \theta}^\ast + \tilde{\bm \theta}$. Then
\begin{align}
  \hat{\bm \theta} & - {\bm \theta}^\ast = \hat{\bm \theta}_{\mathcal M} - {\bm \theta}^\ast + \hat{\bm \theta}_{{\mathcal M}^\perp} 
	= \tilde{\bm \theta}_{\mathcal M} + \tilde{\bm \theta}_{{\mathcal M}^\perp}  \, . \label{aeqn732}
\end{align}
By \cite[Lemma 1]{Negahban2012},
\begin{align}
  {\cal R}( \tilde{\bm \theta}_{{\mathcal M}^\perp} ) \; \le \; & 
	 3 {\cal R}( \tilde{\bm \theta}_{{\mathcal M}} ) 
	  + 4  {\cal R}(  {\bm \theta}^\ast_{{\mathcal M}^\perp}) \, , \label{aeqn734}
\end{align}
if 
\begin{align}
  \lambda_n \ge & 2 {\cal R}^\ast(\nabla {\cal L}({\bm \theta}^\ast))  \, . \label{aeqn736}
\end{align}
Since in our case ${\bm \theta}^\ast_{{\mathcal M}^\perp} = {\bm 0}$, we have ${\cal R}(  {\bm \theta}^\ast_{{\mathcal M}^\perp})=0$. \\
{\bf Lemma 5}.  Under (\ref{eqn200}) and using the notation of Appendix \ref{append1}, 
\begin{align}
  & {\cal R}^\ast(\nabla {\cal L}({\bm \theta}^\ast)) \; \le \; 
	(\epsilon^2+2 M \epsilon) s_n \, \max_{t=1, \cdots, p^2} \| {\bm \theta}_{Gt}^\ast \|_2
	  + 2 \epsilon \,  \quad \bullet \nonumber 
\end{align}
{\it Proof}. Using (\ref{aeqn400}), (\ref{aeqn530}) and (\ref{aeqn722}), we have
\begin{align}
  \nabla {\cal L}({\bm \theta}^\ast) = & {\bm \Delta}_\Gamma {\bm \theta}^\ast +
	        \mbox{bvec}({\bm \Delta}_y) - \mbox{bvec}({\bm \Delta}_x) \, .
	\label{aeqn740}
\end{align}
Expressing it group-wise, with groups $t$ and $t_1$ corresponding to edges $\{j,k\}$ and $\{\ell,q\}$, respectively,
\begin{align}
  (\nabla {\cal L}({\bm \theta}^\ast))_{Gt_1} = & \sum_{t=1}^{p^2} ({\bm \Delta}_\Gamma)_{Gt_1,Gt} {\bm \theta}^\ast_{Gt} +
	        \mbox{bvec}({\bm \Delta}_y)_{Gt_1} \nonumber \\
					& \quad - \mbox{bvec}({\bm \Delta}_x)_{Gt_1} \, .
	\label{aeqn742}
\end{align}
Therefore, by the Cauchy-Schwartz inequality, and using (\ref{aeqn401}), (\ref{aeqn403}) and (\ref{aeqn428}), we have
\begin{align}
 & \|(\nabla {\cal L}({\bm \theta}^\ast))_{Gt_1}\|_2 \le  
        \sum_{t=1}^{p^2} \|({\bm \Delta}_\Gamma)_{Gt_1,Gt}\|_F \, \|{\bm \theta}^\ast_{Gt}\|_2 \nonumber \\
		& \quad\quad\quad +
	        \|\mbox{bvec}({\bm \Delta}_y)_{Gt_1}\|_2 + \| \mbox{bvec}({\bm \Delta}_x)_{Gt_1} \|_2 \nonumber \\
	& \quad \le  \| \bm{\mathcal C}({\bm \Delta}_\Gamma )\|_\infty \, \sum_{t=1}^{p^2} \|{\bm \theta}^\ast_{Gt}\|_2 + 
	  \|{\bm \Delta}_y^{(\ell q)} \|_\infty + \|{\bm \Delta}_y^{(\ell q)} \|_\infty  \nonumber \\
	& \quad \le (\epsilon^2 + 2 M \epsilon) s_n \, \max_{t=1, \cdots, p^2} \| {\bm \theta}_{Gt}^\ast \|_2 + \epsilon + \epsilon \, .
	\label{aeqn744}
\end{align}
By (\ref{aeqn714}) and (\ref{aeqn744}) we have the desired result. $\quad \Box$

{\bf Lemma 6}.  Under (\ref{eqn200}) and the notation of Appendix \ref{append1}, 
\begin{align}
   \delta {\cal L}(\tilde{\bm \theta}, {\bm \theta}^\ast)   \ge \kappa_{\cal L} \, \| \tilde{\bm \theta} \|_2^2
	        \label{aeqn749}
\end{align}
where $\kappa_{\cal L} = \frac{1}{2}\phi_{min}^\ast 
  - 8 s_n (\epsilon^2 + 2 M \epsilon)  $. $\quad \bullet$ \\
{\it Proof}. We have
\begin{align}
 & \tilde{\bm \theta}^\top  (\hat{\bm \Gamma}-{\bm \Gamma}^\ast) \tilde{\bm \theta} =
    \sum_{t_1=1}^{p^2} \sum_{t_2=1}^{p^2} \tilde{\bm \theta}^\top_{Gt_1} ({\bm \Delta}_\Gamma)_{Gt_1,Gt_2} \tilde{\bm \theta}_{Gt_2}
  \, . \label{aeqn750}
\end{align}
Therefore,
\begin{align}
 & |\tilde{\bm \theta}^\top  (\hat{\bm \Gamma}-{\bm \Gamma}^\ast) \tilde{\bm \theta}| \le
    \sum_{t_1=1}^{p^2} \sum_{t_2=1}^{p^2} |\tilde{\bm \theta}^\top_{Gt_1} ({\bm \Delta}_\Gamma)_{Gt_1,Gt_2} \tilde{\bm \theta}_{Gt_2}| \nonumber \\
	& \quad  \le \sum_{t_1=1}^{p^2} \sum_{t_2=1}^{p^2} \|\tilde{\bm \theta}_{Gt_1}\|_2 \,
	  \|({\bm \Delta}_\Gamma)_{Gt_1,Gt_2} \|_F \, \|\tilde{\bm \theta}_{Gt_2}\|_2  \nonumber \\
	& \quad  \le \| \bm{\mathcal C}({\bm \Delta}_\Gamma )\|_\infty \sum_{t_1=1}^{p^2} \sum_{t_2=1}^{p^2} \|\tilde{\bm \theta}_{Gt_1}\|_2 \,
	  \|\tilde{\bm \theta}_{Gt_2}\|_2  \nonumber \\
	& \quad  \le (\epsilon^2 + 2 M \epsilon) \, \|\tilde{\bm \theta}\|_{\bar{\cal G},2}^2 
      \; ,  \label{aeqn752}
\end{align}
where we used (\ref{aeqn705}). We have
\begin{align}
 & \|\tilde{\bm \theta}\|_{\bar{\cal G},2}^2 
  = \| \tilde{\bm \theta}_{\mathcal M} + \tilde{\bm \theta}_{{\mathcal M}^\perp} \|_{\bar{\cal G},2}^2
	= (\| \tilde{\bm \theta}_{\mathcal M} \|_{\bar{\cal G},2} + \|\tilde{\bm \theta}_{{\mathcal M}^\perp} \|_{\bar{\cal G},2})^2 \nonumber \\
	& \quad \overset{(\ref{aeqn734})}{\le} 16 \, \| \tilde{\bm \theta}_{\mathcal M} \|_{\bar{\cal G},2}^2
	   \overset{(\ref{aeqn716})}{\le} 16 s_n \| \tilde{\bm \theta}_{\mathcal M} \|_2^2 \le 16 s_n \| \tilde{\bm \theta} \|_2^2
  \, . \label{aeqn754}
\end{align}
Noting that $\tilde{\bm \theta}^\top  {\bm \Gamma}^\ast) \tilde{\bm \theta}  \ge \phi_{min}^\ast \, \| \tilde{\bm \theta} \|_2^2$
and using (\ref{aeqn726}), (\ref{aeqn752}) and (\ref{aeqn754}), we have
\begin{align}
 & \delta {\cal L}(\tilde{\bm \theta}, {\bm \theta}^\ast)   \ge \big( \frac{1}{2}\phi_{min}^\ast 
  - 8 s_n (\epsilon^2 + 2 M \epsilon) \big) \| \tilde{\bm \theta} \|_2^2 = \kappa_{\cal L} \, \| \tilde{\bm \theta} \|_2^2
  \, , \nonumber 
\end{align}
proving the desired result. $\quad \Box$

{\it Proof of Theorem 2}. First choose $\epsilon$ to make $\kappa_{\cal L} > 0$ in Lemma 6. For instance, suppose we take $8 s_n (\epsilon^2 + 2 M \epsilon) \le \phi_{min}^\ast/4$. Then $\kappa_{\cal L} > \phi_{min}^\ast/4$. Now pick 
\begin{align}
 \epsilon = & C_0 \sqrt{ln(p_n)/n} \; \le \; \min \big\{ M, \frac{\phi_{min}^\ast}{96 s_n M} \big\}
  \, ,  \label{aeqn770}
\end{align}
leading to $8 s_n (\epsilon^2 + 2 M \epsilon) \le 24 s_n M \epsilon \le \phi_{min}^\ast/4$. These upper bounds can be ensured by picking appropriate lower bounds to sample size $n$ and invoking Lemma 2. The choice of $n$ specified in (\ref{eqn452}) satisfies (\ref{aeqn770}) with probability $> 1- 2/p_n^{\tau -2}$. Using $\epsilon = C_0 \sqrt{ln(p_n)/n} \le M$, the lower bound on $\lambda_n$ given in (\ref{eqn450}) satisfies (\ref{aeqn736}) with ${\cal R}^\ast(\nabla {\cal L}({\bm \theta}^\ast))$ as in Lemma 5. By \cite[Theorem 1]{Negahban2012}, $\hat{\bm \theta}$ given by (\ref{aeqn730}) satisfies 
\begin{align}
 \|\hat{\bm \theta}- {\bm \theta}^\ast \|_2 \, \le \, & \frac{3 \lambda_n}{\kappa_{\cal L}} \Psi({\mathcal M})
  \, .  \label{aeqn772}
\end{align}
The left-side of (\ref{aeqn772}) equals $\| \hat{\bm \Delta} - {\bm \Delta}^\ast \|_F$ while the right-side of (\ref{aeqn772}) equals right-side of (\ref{eqn454}) using $\Psi({\mathcal M}) = \sqrt{s_n}$, $\kappa_{\cal L} > \phi_{min}^\ast/4$, and noting that $\max_{t=1, \cdots, p^2} \| {\bm \theta}_{Gt}^\ast \|_2 = \max_{\{k, \ell \} \in V \times V} \| ({\bm \Delta}^\ast)^{(k \ell)} \|_F$. This proves Theorem 2. $\quad \Box$

\bibliographystyle{unsrt}

\end{document}